\documentclass[final]{cvpr}

\usepackage{times}
\usepackage{epsfig}
\usepackage{graphicx}
\usepackage{amsmath}
\usepackage{amssymb}
\usepackage{multicol}
\usepackage{multirow}
\usepackage{makecell}

\usepackage[utf8]{inputenc}
\usepackage{listings}
\usepackage{xcolor}

\definecolor{codegreen}{rgb}{0,0.6,0}
\definecolor{codegray}{rgb}{0.5,0.5,0.5}
\definecolor{codepurple}{rgb}{0.58,0,0.82}
\definecolor{backcolour}{rgb}{0.95,0.95,0.92}

\lstdefinestyle{mystyle}{
    backgroundcolor=\color{backcolour},   
    commentstyle=\color{codegreen},
    keywordstyle=\color{magenta},
    numberstyle=\tiny\color{codegray},
    stringstyle=\color{codepurple},
    basicstyle=\ttfamily\footnotesize,
    breakatwhitespace=false,         
    breaklines=true,                 
    captionpos=b,                    
    keepspaces=true,                 
    showspaces=false,                
    showstringspaces=false,
    showtabs=false,                  
    tabsize=2
}

\usepackage[pagebackref=true,breaklinks=true,colorlinks,bookmarks=false]{hyperref}

\begin{document}
\title{Few-Shot Classification with Feature Map Reconstruction Networks}

\author{Davis Wertheimer\thanks{Equal contribution}\qquad Luming Tang\footnotemark[1]\qquad Bharath Hariharan\\
Cornell University\\
{\tt\small \{dww78,lt453,bh497\}@cornell.edu}
}

\maketitle

\begin{abstract}
In this paper we reformulate few-shot classification as a reconstruction problem in latent space. The ability of the network to reconstruct a query feature map from support features of a given class predicts membership of the query in that class. We introduce a novel mechanism for few-shot classification by regressing directly from support features to query features in closed form, without introducing any new modules or large-scale learnable parameters. The resulting Feature Map Reconstruction Networks are both more performant and computationally efficient than previous approaches. We demonstrate consistent and substantial accuracy gains on four fine-grained benchmarks with varying neural architectures. Our model is also competitive on the non-fine-grained mini-ImageNet and tiered-ImageNet benchmarks with minimal bells and whistles.\footnote{Code is available at {\scriptsize\url{https://github.com/Tsingularity/FRN}}}
\end{abstract}

\section{Introduction}
    Convolutional neural classifiers have achieved excellent performance in a wide range of settings and benchmarks, but this performance is achieved through large quantities of labeled images from the relevant classes. 
    In practice, such a large quantity of human-annotated images may not always be available for the categories of interest. 
    Instances of relevant classes may be rare in the wild, and identifying them may require expensive expert annotators, limiting the availability of training points and labels respectively. 
    These problems are compounded in settings such as robotics, where a model may need to learn and adapt quickly in deployment, without waiting for offline data collection. 
    Producing a performant classifier in these settings requires a neural network that can rapidly fit novel, possibly unseen classes from a small number of reference images. 
    
    \begin{figure}
        \centering
        \includegraphics[width=.95\linewidth]{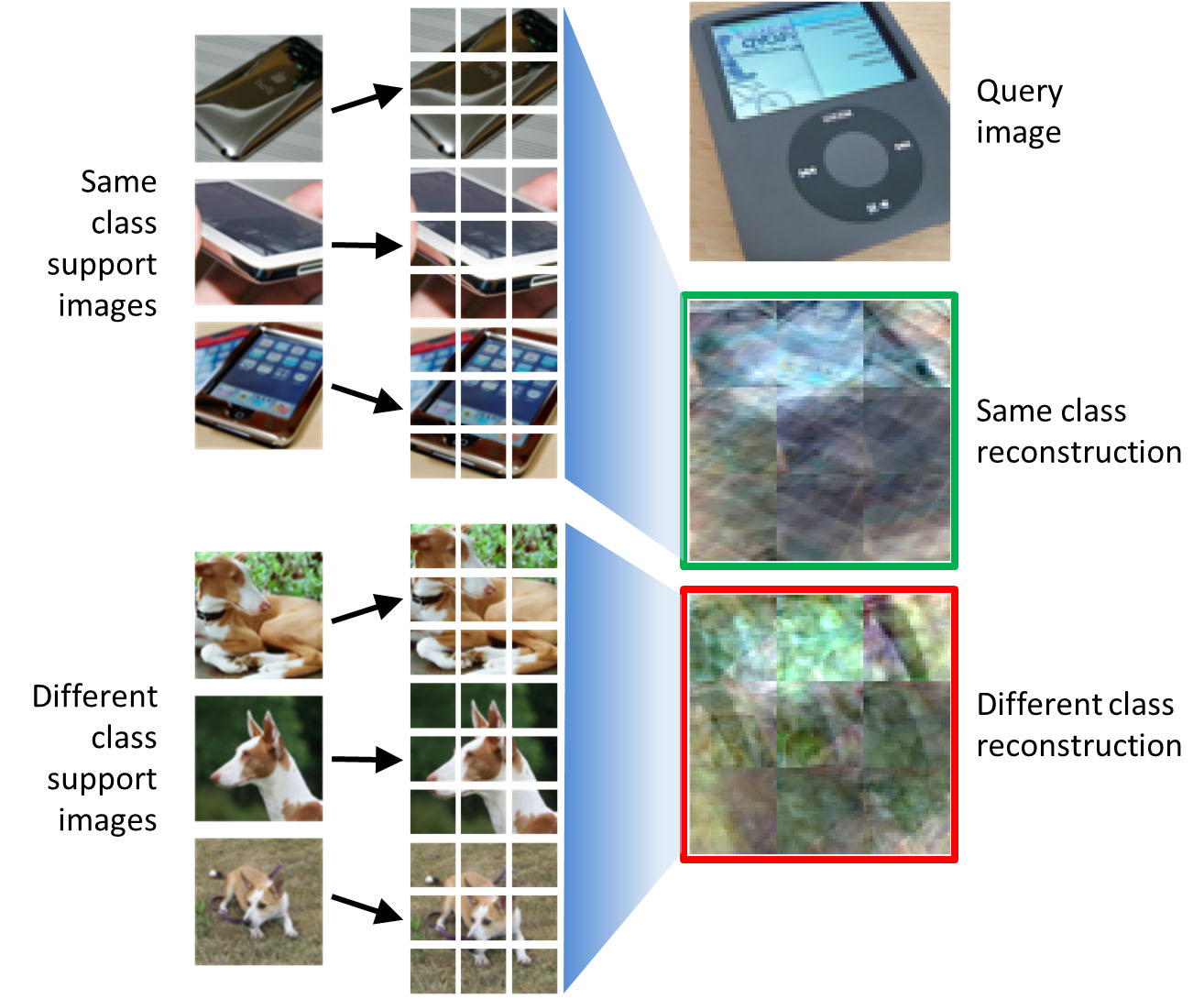}
        \caption{Visual intuition for FRN: we reconstruct each query image as a weighted sum of components from the support images. Reconstructions from the same class are better than reconstructions from different classes, enabling classification. FRN performs the reconstruction in latent space, as opposed to image space, here.}
        \label{fig:pg1}
    \end{figure}

    A promising approach to this problem of \textit{few-shot classification} is the family of \textit{metric learning} techniques, where the standard parametric linear classifier head is replaced with a class-agnostic distance function. 
    Class membership is determined by distance in latent space from a point or points known to belong to each class. 
    Simple distance functions such as cosine~\cite{gidaris2018dynamic,chen2020new} and Euclidean distance~\cite{snell2017prototypical} lead to surprisingly powerful classifiers, though more complex~\cite{simon2020adaptive}, non-Euclidean~\cite{khrulkov2020hyperbolic}, and even learned parametric options~\cite{sung2018learning} are possible, and yield sizable gains. 
    
    One overarching problem common to all these techniques is the fact that the convolutional feature extractors used to learn the metric spaces produce \emph{feature maps} characterizing appearance at a \emph{grid of spatial locations}, whereas the chosen distance functions require a \emph{single vectorial representation for the entire image}. 
    The researcher must decide how to convert the feature map into a vector representation. 
    Optimally, this conversion would preserve the spatial granularity and detail of the feature map without overfitting to pose, but existing, widely-employed approaches do not accomplish this. 
    Global average-pooling, the standard solution for parametric softmax classifiers, averages information from disparate parts of the image, completely discarding spatial details that might be necessary for fine distinctions.  
    Flattening the feature map into a single long vector preserves the individual features~\cite{snell2017prototypical,sung2018learning}, but also encodes the explicit \emph{location} of each feature. 
    This sensitivity to feature location and arrangement (i.e., object pose), regardless of underlying semantic content, is highly undesirable. 
    Larger and more responsive receptive fields will reduce this sensitivity, but instead overfit to specious cues~\cite{doersch2020crosstransformers}.
    We aim to avoid these tradeoffs entirely, preserving spatial detail while disentangling it from location. 
    
    We introduce Feature Map Reconstruction Networks (FRN), which accomplish this by framing class membership as a problem of \emph{reconstructing feature maps}.
    Given a set of images all belonging to a single class, we produce the associated feature maps and collect the component feature vectors \emph{across locations and images} into a single pool of support features. 
    For each query image, we then attempt to reconstruct \emph{every location} in the feature map as a weighted sum of support features, and the negative average squared reconstruction error is used as the class score. 
    Images from the same class should be easier to reconstruct, since their feature maps contain similar embeddings, while images from different classes will be more difficult and produce larger reconstruction errors.
    By evaluating the reconstruction of the full feature map, FRN preserves the spatial details of appearance.
    But by allowing this reconstruction to use feature vectors from \emph{any location} in the support images, FRN explicitly discards nuisance location information.
    
    While prior methods based on feature map reconstruction exist, these methods either rely on constrained iterative procedures~\cite{Zhang_2020_CVPR} or large learned attention modules~\cite{doersch2020crosstransformers,hou2019cross}.
    Instead, we frame feature map reconstruction as a ridge regression problem, allowing us to rapidly calculate a solution in closed form with only a single learned, soft constraint. 
    
    The resulting reconstructions are discriminative and semantically rich, making FRN both simpler and more powerful than prior reconstruction-based approaches. 
    We validate these claims by demonstrating across-the-board superiority on four fine-grained few-shot classification datasets (CUB~\cite{WahCUB_200_2011}, Aircraft~\cite{maji13fine-grained}, meta-iNat and tiered meta-iNat~\cite{wertheimer2019few}) and two general few-shot recognition benchmarks (mini-ImageNet~\cite{vinyals2016matching} and tiered-ImageNet~\cite{ren18fewshotssl}).  These results hold for both shallow and deep network architectures (Conv-4~\cite{snell2017prototypical,lee2019meta} and ResNet-12~\cite{he2016deep,lee2019meta}).

\section{Background and Related Work}

    \textbf{The few-shot learning setup:} Typical few-shot training and evaluation involves sampling task \textit{episodes} from an overarching task distribution -- typically, by repeatedly selecting small subsets from a larger set of classes. 
    Images from each class in the episode are partitioned into a small \textit{support set} and a larger \textit{query set}. 
    The number of classes per episode is referred to as the \textit{way}, while the number of support images per class is the \textit{shot}, so that episodes with five classes and one labeled image per class form a ``5-way, 1-shot'' classification problem. 
    Few-shot classifiers are trained on a large, disjoint set of classes with many labeled images, typically using this same episodic scheme for each batched iteration of SGD. 
    Optimizing the few-shot classifier over the task distribution teaches it to generalize to new tasks from a similar distribution. The classifier learns to learn new tasks, thus episodic few-shot training falls under the umbrella of ``meta-learning'' or ``meta-training''. 
    
   \textbf{Prior work in few-shot learning:} Existing approaches to few-shot learning can be loosely organized into the following two main-stream families. Optimization-based methods~\cite{finn2017model,rusu2018meta,nichol2018first} aim to learn a good parameter initialization for the classifier. These learned weights can then be quickly adapted to novel classes using gradient-based optimization on only a few labeled samples. Metric-based methods, on the other hand, aim to learn a task-independent embedding that can generalize to novel categories under a chosen distance metric, such as Euclidean distance~\cite{snell2017prototypical}, cosine distance~\cite{gidaris2018dynamic}, hyperbolic distance~\cite{khrulkov2020hyperbolic}, or a distance parameterized by a neural network~\cite{sung2018learning}. 
    
    As an alternative to the standard meta-learning framework, many recent papers~\cite{chen2019closerfewshot,tian2020rethinking,wang2019simpleshot} study the performance of standard end-to-end pre-trained classifiers on few-shot tasks. Given minimal modification, these classifiers are actually competitive with or even outperform episodic meta-training methods. Therefore some recent works~\cite{Zhang_2020_CVPR,ye2020fewshot,chen2020new} take advantage of both, and utilize meta-learning after pre-training, further boosting performance.
    
    \textbf{Few-shot classification through reconstruction:} 
    Feature reconstruction is a classic approach~\cite{LK20yearson} to object tracking and alignment~\cite{dollar10pose,cao12faces,sun15hands,wang18deeplk}, but has only recently been utilized for few-shot classification.  
    DeepEMD~\cite{Zhang_2020_CVPR} formulates reconstruction as an optimal transport problem.
    This formulation is sophisticated and powerful, but training and inference come with significant computational cost, due to the reliance on iterative constrained convex optimization solvers and test-time SGD. 
    CrossTransformer~\cite{doersch2020crosstransformers} and CrossAttention~\cite{hou2019cross} add attention modules that project query features into the space of support features (or vice versa), and compare the class-conditioned projections to the target to predict class membership. 
    These attention-based approaches introduce many additional learned parameters over and above the network backbone, and place substantial constraints on the projection matrix (weights are non-negative and rows must sum to 1). 
    In contrast, FRN efficiently calculates minimally constrained, least-squares-optimal reconstructions in closed form.

    \textbf{Closed-form solvers in few-shot learning:} The use of closed-form solvers for few-shot classification is also not entirely new, though to our knowledge they have not been applied in the explicit context of feature reconstruction. 
    ~\cite{bertinetto2018metalearning} uses ridge regression to map features directly to classification labels, while~\cite{sung2018learning} accomplishes the same mapping with differentiable SVMs. 
    Deep Subspace Networks~\cite{simon2020adaptive} use the closed-formed projection distance from query embeddings to subspaces spanned by support points as the similarity measure. 
    In contrast, FRN uses closed-form ridge regression to reconstruct entire feature maps, rather than performing direct comparisons between single points in latent space, or regressing directly to class label targets.
    
\section{Method}
\begin{figure}
        \centering
        \includegraphics[width=.95\linewidth]{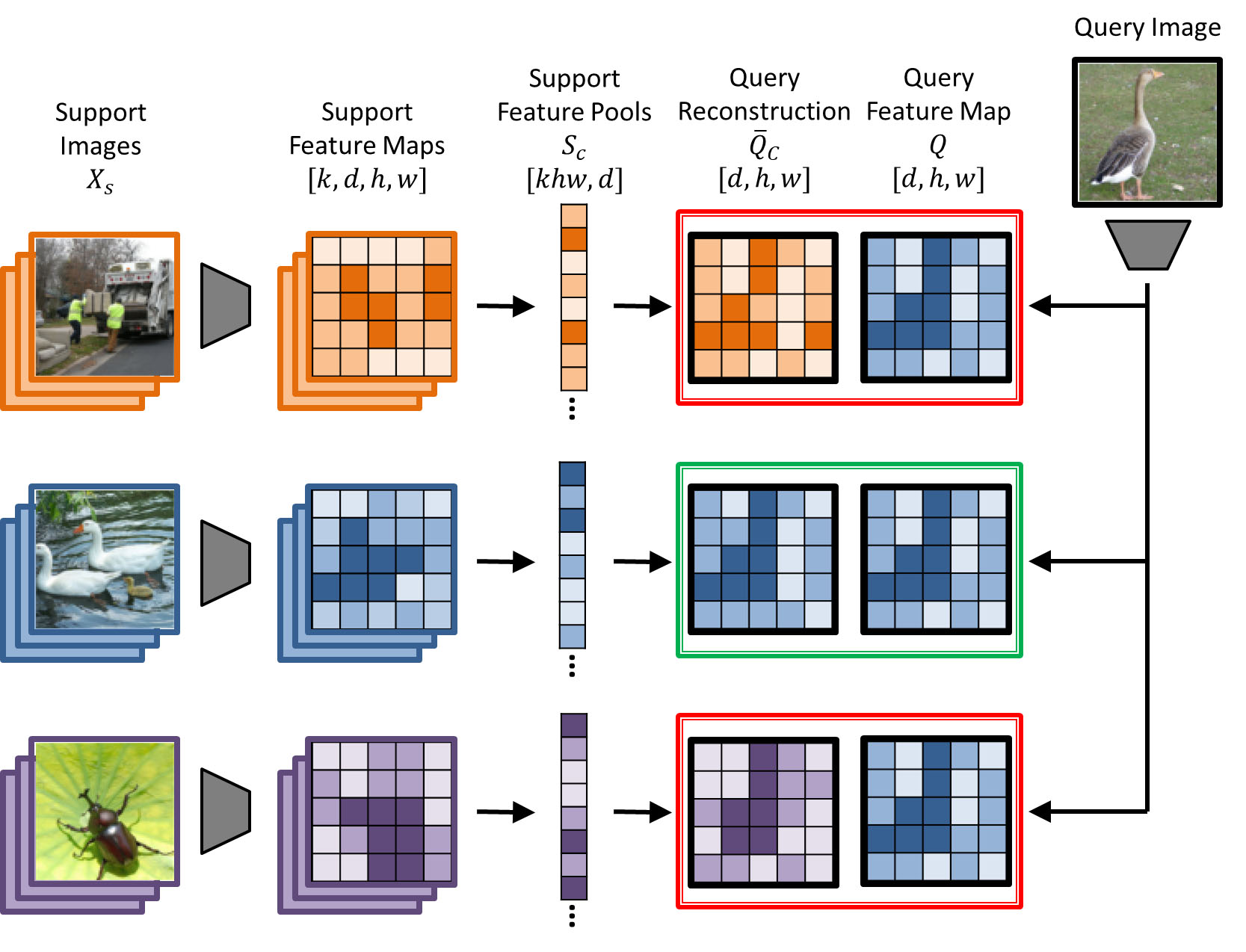}
        \caption{Overview of FRN classification for a k-shot problem. Support images are converted into feature maps (left), which are aggregated into class-conditional pools (middle). The best-fit reconstruction of the query feature map is calculated for each category, and the closest candidate yields the predicted class (right). $h,w$ is feature map resolution and $d$ is the number of channels.}
        \label{fig:method}
    \end{figure}

    Feature Map Reconstruction Networks use the quality of query feature map reconstructions from support features as a proxy for class membership. The pool of features associated with each class in the episode is used to calculate a candidate reconstruction, with a better reconstruction indicating higher confidence for the associated class. 
    In this section we describe the reconstruction mechanism of FRN in detail, and derive the closed-form solution used to calculate the reconstruction error and resulting class score. An overview is provided in Fig.~\ref{fig:method}. We discuss memory-efficient implementations and an optional pre-training scheme, and draw comparisons to prior reconstruction-based approaches. 
    
    \subsection{Feature Map Ridge Regression}
    Let $X_s$ denote the set of support images with corresponding class labels in an $n$-way, $k$-shot episode. We wish to predict a class label $y_q$ for a single input query image $x_q$. 
    
    The output of the convolutional feature extractor for $x_q$ is a feature map $Q\in \mathbb{R}^{r\times d}$, with $r$ the spatial resolution (height times width) of the feature map, and $d$ the number of channels. 
    For each class $c\in C$, we pool all features from the $k$ support images into a single matrix of support features $S_c\in \mathbb{R}^{kr\times d}$. 
    We then attempt to reconstruct $Q$ as a weighted sum of rows in $S_c$ by finding the matrix $W\in \mathbb{R}^{r\times kr}$ such that $WS_c\approx Q$. Finding the optimal $\bar W$ amounts to solving the linear least-squares problem:
    \begin{align}
    \label{eq:linprogram}
        \bar W = \underset{W}{\text{arg min\ \ }} ||Q-WS_c||^2 + \lambda ||W||^2
    \end{align}
    where $|| \cdot ||$ is the Frobenius norm and $\lambda$ weights the ridge regression penalty term used to ensure tractability when the linear system is over- or under-constrained ($kr\ne d$).
    
    The foremost benefit of the ridge regression formulation is that it admits a widely-known closed-form solution for $\bar W$ and the optimal reconstruction $\bar Q_c$ as follows:
    \begin{align}
    \label{q_origin}
        \bar W &= QS_c^T(S_cS_c^T+\lambda I)^{-1}\\
        \bar Q_c &= \bar WS_c
    \end{align}
    For a given class $c$, the negative mean squared Euclidean distance between $Q$ and $\bar Q_c$ over all feature map locations yields the scalar probability logit.
    We also incorporate a learnable temperature factor $\gamma$, following
    \cite{chen2020new,gidaris2018dynamic,ye2020fewshot}.
    The final predicted probability is thus given by:
    \begin{align}
        \langle Q,\bar Q_c \rangle &= \frac{1}{r}||Q-\bar Q_c||^2\\
        P(y_q=c|x_q) &= \frac{e^{(-\gamma\langle Q,\bar Q_c \rangle)} }
        {\sum_{c'\in C} e^{(-\gamma\langle Q,\bar Q_{c'} \rangle)} }
    \end{align}
    We optimize our network by sending the predicted class probabilities for the query images in each episode through a cross-entropy loss, as in standard episodic meta-training.
    An overview of this process can be found in Fig.~\ref{fig:method}.

    \subsection{Learning the Degree of Regularization}
    The difficulty of the Eq. \ref{eq:linprogram} reconstruction problem varies widely.
    If $kr > d$, reconstruction may become trivial, as the support features can span the feature space. Conversely, reconstruction is difficult when $d> kr$. 
    To ensure a balanced objective and stable training, we therefore rescale the regularizer $\lambda$ by $\frac{kr}{d}$.
    This has the added benefit of making our model somewhat robust to shot, in that concatenating a support pool to itself now yields unchanged reconstructions.
    
    Even with rescaling, though, it is not immediately clear how one should set the regularizer $\lambda$.
    Instead of choosing heuristically, we have the network \emph{learn} $\lambda$ through meta-learning. This is significant, as it allows the network to pick a degree of regularization such that reconstruction is \emph{discriminative}, rather than strictly least-squares optimal. 
    
    Changing $\lambda$ can have multiple effects.
    Large $\lambda$ discourages overreliance on particular weights in $W$, but also reduces the norm of the reconstruction, increasing reconstruction error and limiting discriminative power.
    We therefore disentangle the degree of regularization from the magnitude of $\bar Q_c$ by introducing a \emph{learned recalibration term} $\rho$: 
    \begin{align}
        \bar Q_c = \rho \bar W S_c
    \end{align} 
    By increasing $\rho$ alongside $\lambda$, the network gains the ability to penalize large weights without sending all reconstructions to the origin at the same time. 
    $\lambda$ and $\rho$ are parameterized as $e^\alpha$ and $e^\beta$ to ensure non-negativity, with $\alpha$ and $\beta$ initialized to zero.
    Thus, all together, our final prediction is given by:
    \begin{align}
    \lambda = \frac{kr}{d}e^\alpha \quad&\quad\quad\quad  \rho = e^\beta \\
    \label{eq:q_full}
        \bar Q_c = \rho \bar W S_c &= \rho QS_c^T(S_cS_c^T+\lambda I)^{-1}S_c\\
    P(y_q=c|x_q) &= \frac{e^{(-\gamma\langle Q,\bar Q_c \rangle)} }
        {\sum_{c'\in C} e^{(-\gamma\langle Q,\bar Q_{c'} \rangle)} }    
    \end{align}
    The model is meta-trained in a similar manner to prior work: sample episodes from a labeled base class dataset and minimize cross entropy on the predicted query labels~\cite{snell2017prototypical}.
    
    Our approach introduces only three learned parameters: $\alpha,\beta$ and $\gamma$. The temperature $\gamma$ appears in prior work~\cite{gidaris2018dynamic,chen2020new,ye2020fewshot}. Ablations on $\alpha$ and $\beta$ can be found in Sec.~\ref{sec:ablation}.

    \subsection{Parallelization}
    While we have described our approach as finding reconstructions for a single query image, it is relatively straightforward to find the reconstructions for an entire batch of query images. 
    We have already calculated the optimal reconstruction for each of the $r$ feature vectors in $Q$ independently; all we need to do for a batch of $b$ images is to pool the features into a larger matrix $Q'\in \mathbb{R}^{br\times d}$ and run the algorithm as written. 
    Thus for an $n$-way episode we will only ever need to run the algorithm $n$ times, once for each support matrix $S_c$, regardless of the quantity or arrangement of queries. These $n$ runs can also be parallelized, given parallel implementations of matrix multiplication and inversion.
    
    \subsection{Alternative Formulation}
    \label{sec:alter}
    The formula for $\bar Q$ in Eq.~\ref{eq:q_full} is efficient to compute when $d> kr$, as the most expensive step is inverting a $kr\times kr$ matrix that does not grow with $d$. Computing the matrix product from left to right also avoids storing a potentially large $d\times d$ matrix in memory. However, if feature maps are large or the shot number is particularly high ($kr> d$), Eq.~\ref{eq:q_full} may quickly become infeasible to compute. In this case an alternative formulation for $\bar Q$ exists, which swaps $d$ for $kr$ in terms of computational requirements. This formulation is owed to the Woodbury Identity~\cite{Petersen2008} as applied in~\cite{bertinetto2018metalearning}:
    \begin{align}
    \label{eq:q_alter}
        \bar Q_c = \rho \bar WS_c = \rho Q(S_c^TS_c+\lambda I)^{-1}S_c^TS_c
    \end{align}
    Here, the most expensive step is a $d\times d$ matrix inversion, and computing the product from right to left avoids storing any large $kr\times kr$ or $br\times kr$ matrices in memory.
    As $r$ and $d$ are determined by the network architecture, the researcher may employ either formulation depending on $k$. The network can also decide on the fly at test time. 
    In terms of classifier performance the two formulations are algebraically equivalent, and pseudo-code for both is provided in Supplementary Materials (SM) Sec.~\ref{sup:code}. For consistency, we employ Eq.~\ref{eq:q_alter} in our implementations.

    \subsection{Auxiliary Loss}
    In addition to the classification loss, we employ an auxiliary loss that encourages support features from different classes to span the latent space~\cite{simon2020adaptive}:
    \begin{align}
    \label{eq:aux}
        L_{\rm aux} = \sum_{i\in C} \sum_{j\in C, j\ne i} ||\hat S_i \hat S_j^T||^2
    \end{align}
    where $\hat S$ is row-normalized, with features projected to the unit sphere. This loss encourages orthogonality between features from different classes. Similar to~\cite{simon2020adaptive}, we downscale this loss by a factor of $0.03$. We use $L_{\rm aux}$ as the auxiliary loss in our subspace network implementation~\cite{simon2020adaptive}, and it replaces the SimCLR episodes in our CrossTransformer implementation~\cite{doersch2020crosstransformers}. We include it in our own model for consistency, and include an ablation study in Sec.~\ref{sec:ablation}.

    \subsection{Pre-Training}
    \label{sec:pretrain}
    Prior work~\cite{chen2020new,ye2020fewshot} has demonstrated that few-shot classifiers can benefit greatly from non-episodic pre-training. For traditional metric learning based approaches, the feature extractor is initially trained as a linear classifier with global average-pooling on the full set of training classes. The linear layer is subsequently discarded, and the feature extractor is fine-tuned episodically. 
    
    This pre-training does not work out-of-the-box for FRN due to its novel classification mechanism.
    Because the linear classifier uses average-pooling, the feature extractor does not learn spatially distinct feature maps in the way FRN requires (see Sec.~\ref{sec:ablation} for analysis).
    
    We therefore devise a new pre-training scheme for FRN.
    To keep the classifier consistent with FRN meta-training, we continue to use feature reconstruction error as the predicted class logit. 
    Similar to~\cite{Zhang_2020_CVPR}, the classification head is parametrized as a set of class-specific \emph{dummy feature maps}, where we introduce a learnable matrix $M_c\in\mathbb{R}^{r\times d}$ for each category $c$, acting as a proxy for $S_c$. Following Eq.~\ref{eq:q_alter}, the prediction for a sample $x_q$ with feature map $Q\in\mathbb{R}^{r\times d}$ is:
    \begin{align}
        \bar Q_c &= \rho Q(M_c^T M_c+\lambda I)^{-1}M_c^T M_c\\
        P(y_q=c|x_q) &= \frac{e^{(-\gamma\langle Q,\bar Q_c \rangle)} }
        {\sum_{c'\in C} e^{(-\gamma\langle Q,\bar Q_{c'} \rangle)} }
    \end{align}
    It should be noted that $C$ in this setting is no longer the sampled subset of episode categories, but rather the entire set of training classes (e.g., $|C|=64$ for mini-ImageNet).
    We then use this output probability distribution to calculate the standard cross-entropy classification loss. During the pre-training stage, we fix $\alpha=\beta=0$ but keep $\gamma$ a learnable parameter. After pre-training is finished, all learned matrices $\{M_c|c\in C\}$ are discarded (similar to the pre-trained MLP classifier in~\cite{tian2020rethinking,ye2020fewshot,wang2019simpleshot,chen2019closerfewshot,chen2020new}). The pre-trained model size is thus the same as when trained from scratch. 
    
    While pre-training is broadly applicable and generally boosts performance, for the sake of fairness we do not pre-train any of our fine-grained experiments, as baseline methods do not consistently pre-train in these settings.

\subsection{Relation to Prior Reconstructive Classifiers}
    
    \textbf{DeepEMD}~\cite{Zhang_2020_CVPR}: FRN uniquely combines feature map comparison with an unconstrained, closed-form reconstruction objective; prior approaches include one or the other, but not both. Like FRN, DeepEMD solves for a $r\times kr$ reconstruction matrix $\bar W$ and uses reconstruction quality (measured as transport cost) as a proxy for class membership. 
    This technique is more sophisticated than ridge regression, but also highly constrained. As a transport matrix, $\bar W$ must hold nonnegative values, with rows and columns that sum to 1. More importantly, $\bar W$ cannot be calculated in closed form, requiring an iterative solver that can be slow in practice and scales poorly to $k$ greater than one.
    We found computing the FRN reconstruction orders of magnitude faster than the EMD equivalent (see Table~\ref{tab:speedsmall}).
    DeepEMD also requires finetuning via back-propagation at test time, whereas our approach scales out of the box to a range of $k,r,d$. 
\begin{table}
\centering
\resizebox{.48\textwidth}{!}{
\scriptsize
\setlength\tabcolsep{5pt}
\hskip-.02\textwidth
\begin{tabular} {  l  c  c  c}
\hline
\textbf{Model} & \textbf{Solver}& \textbf{1-shot Latency} & \textbf{5-shot Latency} \\
\hline
DeepEMD~\cite{Zhang_2020_CVPR} & qpth~\cite{amos2017optnet} & 23,275 & $>$800,000 \\
DeepEMD~\cite{Zhang_2020_CVPR} & OpenCV~\cite{opencv_library} & 178 & 18,292 \\
FRN (ours) & Eq.~\ref{eq:q_alter} & \textbf{73} & \textbf{88}\\
FRN (ours) & Eq.~\ref{eq:q_full} & \textbf{63} & \textbf{79}\\
\hline
\end{tabular}
}
\vspace{.8mm}
\caption{Latency (ms) for 5-way mini-ImageNet evaluation with ResNet-12.
Detailed discussion and comparison in SM Sec.~\ref{sup:speed}.
}
\label{tab:speedsmall}
\end{table}

    \textbf{CrossTransformer (CTX)}~\cite{doersch2020crosstransformers}: CTX and related approaches~\cite{hou2019cross} are more similar to FRN, in that they explicitly produce class-wise linear reconstructions $\bar Q_c = \bar WS_c$. 
    However, rather than solving for $\bar W$, these methods approximate it using attention and extra learned projection layers. 
    CTX reprojects the feature pools $S_c$ and $Q$ into two different ``key" and ``value" subspaces, yielding $S_1,Q_1$ and $S_2,Q_2$. The reconstruction of $Q_2$ is given by:
    \begin{align}
    \label{eq:ctx}
        \bar Q_2 &= \sigma(\frac{1}{\sqrt{d}}Q_1S_1^T)S_2
    \end{align}
    where $\sigma(\cdot)$ denotes a row-wise softmax.
    While Eq.~\ref{eq:ctx} is loosely analogous to Eq.~\ref{eq:q_full}, with the $\sqrt{d}$-scaled softmax replacing the inverted matrix term, we find that performance differs in practice. The CTX layer is also somewhat unwieldy: the two reprojection layers introduce extra parameters into the network, and during training it is necessary to store the $br\times kr$ matrix of attention weights $\sigma(\frac{1}{\sqrt{d}}Q_1S_1^T)$ for back-propagation. This can lead to a noticeable memory footprint as these values increase -- while we did not observe a difference in our experimental settings, simply increasing $r$ from $5\times5$ to $10\times10$ in our implementation was sufficient to introduce a 2-3GB overhead (see SM Sec.~\ref{sup:speed}). 
    
    \textbf{Deep Subspace Networks (DSN)}~\cite{simon2020adaptive}: DSN predicts class membership by calculating the distance between the query point and its projections onto the latent subspaces formed by the support images for each class. This is analogous to our approach with $r=1$, with average-pooling performing the spatial reduction. The crucial difference is that DSN assumes (accurately) that $d> k$, whereas in our setting it is not always the case that $d> kr$. In fact, for many of our models $S$ spans the latent space, so the projection interpretation falls apart and we instead rely on the ridge regression regularizer to keep the problem well-posed. 
    
    \begin{table}
    \centering
    \resizebox{.4\textwidth}{!}{
    \scriptsize
    \hskip-.01\textwidth
    \begin{tabular}{|c|c|c|}
        \hline
        Model & \makecell{Feature map} & \makecell{Regression objective} \\
        \hline
        Proto~\cite{snell2017prototypical} & $\times$ & $\times$ \\
        DSN~\cite{simon2020adaptive} & $\times$ & \checkmark \\
        CTX~\cite{doersch2020crosstransformers} & \checkmark & $\times$ \\
        FRN (ours) & \checkmark & \checkmark \\
        \hline
    \end{tabular}
    }
    \vspace{.8mm}
    \caption{Relationships between our implemented models.}
    \label{tab:taxonomy}
    \end{table}
    
    Of the methods that produce explicit reconstructions, CTX compares feature maps while DSN utilizes a closed-form regression objective. FRN captures both concepts, leading to the organization shown in Table~\ref{tab:taxonomy}. 
    We thus re-implement CTX and DSN as direct comparison baselines (details in SM Sec.~\ref{sup:comparison}). 
    As shown in the following section, FRN leverages a unique synergy between these concepts to improve even when CTX or DSN on their own do not. 
    
\section{Experiments}
\label{sec:experiments}
    Feature Map Reconstruction Networks focus on spatial details without overfitting to pose, making them particularly powerful in the fine-grained few-shot recognition setting, where details are important and pose is not discriminative. 
    We demonstrate clear superiority on four such benchmarks. 
    For general few-shot learning, FRN with pre-training achieves highly competitive results without extra bells or whistles.
    
    \textbf{Implementation details}: We conduct experiments on two widely used backbones: 4-layer ConvNet (Conv-4) and ResNet-12. 
    Same as~\cite{ye2020fewshot,lee2019meta}, Conv-4 consists of 4 consecutive 64-channel convolution blocks that each downsample by a factor of 2. 
    The shape of the output feature maps for input images of size 84$\times$84 is thus 64$\times$5$\times$5. For ResNet-12, we use the same implementation as~\cite{ye2020fewshot,tian2020rethinking,lee2019meta}. 
    The input image size is the same as Conv-4 and the output feature map shape is 640$\times$5$\times$5. During training, we use the standard data augmentation as in~\cite{ye2020fewshot,Zhang_2020_CVPR,wang2019simpleshot,chen2019closerfewshot}, which includes random crop, right-left flip and color jitter. 
    Further training details can be found in SM Sec.~\ref{sup:training}.
    
    Evaluation is performed on standard 5-way, 1-shot and 5-shot settings. Accuracy scores and 95\% confidence intervals are obtained over 10,000 trials, as in \cite{ye2020fewshot,chen2020new,wang2019simpleshot}.

\subsection{Fine-Grained Few-Shot Classification}
\label{sec:fine_grained}

    For our fine-grained experiments, we re-implement three baselines: Prototypical Networks (ProtoNet$^{\dag}$)~\cite{snell2017prototypical}, CTX$^{\dag}$~\cite{doersch2020crosstransformers}, and DSN$^{\dag}$~\cite{simon2020adaptive}, where $\dag$ denotes our implementation. 
    For fair comparison, we do not use pre-training for any of our implemented models here,
    or tune FRN hyperparameters separately from baseline models.

    \textbf{CUB}~\cite{WahCUB_200_2011} consists of 11,788 images from 200 bird classes. Following~\cite{chen2019closerfewshot}, we randomly split categories into 100 classes for training, 50 for validation and 50 for evaluation. Our split is identical to ~\cite{tang2020revisiting} (discussion of class splits can be found in SM Sec.~\ref{sup:cub}). Prior work on this benchmark pre-processes the data in different ways:~\cite{chen2019closerfewshot} uses raw images as input, while~\cite{ye2020fewshot,Zhang_2020_CVPR} crop each image to a human-annotated bounding box. 
    We experiment on both settings for fair comparison. 
    
        \begin{table}
        \centering
        \resizebox{.45\textwidth}{!}{
        \hskip-.01\textwidth
        \begin{tabular} {  l  c c c c}
        \hline
        \-\ & \multicolumn{2}{c}{\textbf{Conv-4}} & \multicolumn{2}{c}{\textbf{ResNet-12}} \\
        \textbf{Model} & \textbf{1-shot} & \textbf{5-shot} & \textbf{1-shot} & \textbf{5-shot}\\
        \hline
        MatchNet$^{\flat}$~\cite{vinyals2016matching,ye2020fewshot, Zhang_2020_CVPR} &	67.73 &	79.00 &	71.87 & 85.08\\
        ProtoNet$^{\flat}$~\cite{snell2017prototypical,ye2020fewshot, Zhang_2020_CVPR} &	63.73	& 81.50&	66.09&	82.50 \\
        Hyperbolic~\cite{khrulkov2020hyperbolic}& 64.02 & 82.53 & - & - \\
        FEAT$^{\flat}$~\cite{ye2020fewshot} & 68.87 &82.90 & - & - \\
        DeepEMD$^{\flat}$~\cite{Zhang_2020_CVPR} & - & - & 75.65	& 88.69 \\
        \hline
        ProtoNet$^{\dag}$~\cite{snell2017prototypical}& 63.21 &	83.88 & 79.09 & 90.59 \\
        DSN$^{\dag}$~\cite{simon2020adaptive}&66.01	&85.41	&80.80	&91.19\\
        CTX$^{\dag}$~\cite{doersch2020crosstransformers} & 69.64	& 87.31	& 78.47	& 90.90\\
        \hline
        FRN (ours) & \textbf{73.48} & \textbf{88.43} & \textbf{83.16} & \textbf{92.59} \\
        \hline
        \end{tabular}
        }
        \vspace{.8mm}
        \caption{Performance on CUB using bounding-box cropped images as input.
        $\flat$: use of non-episodic pre-training. 
        Confidence intervals for our implemented models are all below 0.24.}
        \label{tab:cub_crop}
        \end{table}
        
        \begin{table}
        \centering
        \resizebox{.45\textwidth}{!}{
        \hskip-.02\textwidth
        \begin{tabular} {  l  c  c  c}
        \hline
        \textbf{Model} & \textbf{Backbone} & \textbf{1-shot} & \textbf{5-shot} \\
        \hline
        Baseline$^{\flat}$~\cite{chen2019closerfewshot} & ResNet-18 & 65.51$\pm$0.87 & 82.85$\pm$0.55  \\
        Baseline++$^{\flat}$~\cite{chen2019closerfewshot} & ResNet-18 & 67.02$\pm$0.90 & 83.58$\pm$0.54  \\
        MatchNet~\cite{chen2019closerfewshot,vinyals2016matching} &	ResNet-18&	73.49$\pm$0.89	&84.45$\pm$0.58 \\
        ProtoNet~\cite{chen2019closerfewshot,snell2017prototypical} &	ResNet-18&	72.99$\pm$0.88 & 86.64$\pm$0.51\\
        MAML~\cite{chen2019closerfewshot,finn2017model} &	ResNet-18&	68.42$\pm$1.07 &	83.47$\pm$0.62\\
        RelationNet~\cite{chen2019closerfewshot,sung2018learning}&	ResNet-18&	68.58$\pm$0.94 &	84.05$\pm$0.56\\
        S2M2$^{\flat}$~\cite{mangla2020charting} & ResNet-18 & 71.43$\pm$0.28 & 85.55$\pm$0.52\\
        Neg-Cosine$^{\flat}$~\cite{liu2020negative} & ResNet-18 & 72.66$\pm$0.85 & 89.40$\pm$0.43 \\
        Afrasiyabi $\etal^{\flat}$~\cite{Afrasiyabi_2020_ECCV} &ResNet-18& 74.22$\pm$1.09& 88.65$\pm$0.55 \\
        \hline
        ProtoNet$^{\dag}$~\cite{snell2017prototypical}&	ResNet-12& 78.60$\pm$0.22 &89.73$\pm$0.12 \\
        DSN$^{\dag}$~\cite{simon2020adaptive}&ResNet-12&79.96$\pm$0.21 & 91.41$\pm$0.34\\
        CTX$^{\dag}$~\cite{doersch2020crosstransformers} &ResNet-12& 79.34$\pm$0.21 & 91.42$\pm$0.11\\
        \hline
        FRN (ours) &	ResNet-12&	\textbf{83.55$\pm$0.19}&	\textbf{92.92$\pm$0.10}\\
        \hline
        \end{tabular}
        }
        \vspace{.8mm}
        \caption{Performance on CUB using raw images as input.
        $\flat$: use of non-episodic pre-training.
        }
        \label{tab:cub_origin}
        \end{table}

    \textbf{Aircraft}~\cite{maji13fine-grained} contains 10,000 images spanning 100 airplane models. Following the same ratio as CUB, we randomly split classes into 50 train, 25 validation and 25 test. 
    Images are pre-cropped to the provided bounding box. 
        
        \begin{table}
        \centering
        \resizebox{.42\textwidth}{!}{
        \scriptsize
        \hskip-.01\textwidth
        \begin{tabular} {  l  c c c c}
        \hline
        \-\ & \multicolumn{2}{c}{\textbf{Conv-4}} & \multicolumn{2}{c}{\textbf{ResNet-12}} \\
        \textbf{Model} & \textbf{1-shot} & \textbf{5-shot} & \textbf{1-shot} & \textbf{5-shot}\\
        \hline
        ProtoNet$^{\dag}$~\cite{snell2017prototypical}& 47.72 &	69.42 & 66.57&	82.37\\
        DSN$^{\dag}$~\cite{simon2020adaptive}&49.63	& 66.36 & 68.16 &	81.85\\
        CTX$^{\dag}$~\cite{doersch2020crosstransformers} & 49.67	& 69.06 & 65.60	& 80.20\\
        \hline
        FRN (ours) & \textbf{53.20} & \textbf{71.17} & \textbf{70.17} & \textbf{83.81} \\
        \hline
        \end{tabular}
        }
        \vspace{.8mm}
        \caption{Performance on Aircraft.
        All $95\%$ confidence intervals are below 0.25.}
        \label{tab:aircraft}
        \end{table}

    \textbf{meta-iNat}~\cite{wertheimer2019few,van2018inaturalist} is a benchmark of animal species in the wild. This benchmark is particularly difficult, as classes are unbalanced, distinctions are fine-grained, and images are not cropped or centered, and may contain multiple animal instances. We follow the class split proposed by~\cite{wertheimer2019few}: of 1135 classes with between 50 and 1000 images, one fifth (227) are assigned to evaluation and the rest to training. While~\cite{wertheimer2019few} propose a full 227-way, $k$-shot evaluation scheme with $10\le k\le 200$, we instead perform standard 5-way, 1-shot and 5-shot evaluation, and leave extension to higher shots and unbalanced classes for future work. 
    
    \textbf{tiered meta-iNat}~\cite{wertheimer2019few} represents a more difficult version of meta-iNat where a large domain gap is introduced between train and test classes. The 354 test classes are populated by insects and arachnids, while the remaining 781 classes (mammals, birds, reptiles, etc.) form the training set. Training and evaluation are otherwise the same. 
    
    Results on fine-grained benchmarks can be found in Tables~\ref{tab:cub_crop}, \ref{tab:cub_origin}, \ref{tab:aircraft}, and \ref{tab:inat}, corresponding to cropped CUB, uncropped CUB, Aircraft, and combined meta-iNat and tiered meta-iNat, respectively. 
    FRN is superior across the board, with a notable \textbf{2-7 point jump in accuracy} (mean 3.5) from the nearest baseline in all 1-shot settings. 

    Note that our re-implemented baselines in Tables~\ref{tab:cub_crop} and \ref{tab:cub_origin} are competitive with (and in some cases beat outright) prior published numbers. This shows that in the experiments without prior numbers, our baselines still provide fair competition. We do not give FRN an unfair edge -- if anything, our baselines are more competitive, not less.
    
    Based on the above observations, we conclude that FRN is broadly effective at fine-grained few-shot classification. 

        \begin{table}
        \centering
        \resizebox{.45\textwidth}{!}{
        \scriptsize
        \begin{tabular} {  l  c c c c}
        \hline
        \-\ & \multicolumn{2}{c}{\textbf{meta-iNat}} & \multicolumn{2}{c}{\textbf{tiered meta-iNat}} \\
        \textbf{Model} & \textbf{1-shot} & \textbf{5-shot} & \textbf{1-shot} & \textbf{5-shot}\\
        \hline
        ProtoNet$^{\dag}$~\cite{snell2017prototypical}& 55.34 &  76.43 & 34.34	& 57.13	\\
        Covar. pool$^\dag$~\cite{wertheimer2019few}& 57.15 & 77.20 & 36.06 & 57.48\\
        DSN$^{\dag}$~\cite{simon2020adaptive}&58.08 & 77.38 &36.82 &60.11	\\
        CTX$^{\dag}$~\cite{doersch2020crosstransformers} &60.03 & 78.80 &36.83 &60.84	\\
        \hline
        FRN (ours) & \textbf{62.42}	& \textbf{80.45}	& \textbf{43.91} & \textbf{63.36} \\
        \hline
        \end{tabular}
        }
        \vspace{.8mm}
        \caption{Performance on meta-iNat and tiered meta-iNat using Conv-4 backbones.
        All $95\%$ confidence intervals are below 0.24.}
        \label{tab:inat}
        \end{table}

\subsection{General Few-Shot Classification}
\label{sec:general}

    \begin{table*}
    \centering
    \resizebox{.8\textwidth}{!}{
    \scriptsize
    \hskip-.02\textwidth
    \begin{tabular} {  l  c  c  c  c  c}
    \hline
    \-\ & \-\ & \multicolumn{2}{c}{\textbf{mini-ImageNet}} & \multicolumn{2}{c}{\textbf{tiered-ImageNet}}\\
    \textbf{Model} & \textbf{Backbone} & \textbf{1-shot} & \textbf{5-shot} & \textbf{1-shot} & \textbf{5-shot}\\
    \hline
    MatchNet$^{\flat\heartsuit}$~\cite{vinyals2016matching,ye2020fewshot,Zhang_2020_CVPR} & ResNet-12 & {65.64$\pm$0.20} & 78.72$\pm$0.15 & 68.50$\pm$0.92 & 80.60$\pm$0.71\\ 
    ProtoNet$^{\flat}$~\cite{snell2017prototypical,ye2020fewshot}& ResNet-12 & 62.39$\pm$0.21 & 80.53$\pm$0.14 & 68.23$\pm$0.23 & 84.03$\pm$0.16\\ 
    
    MetaOptNet$^{\sharp}$~\cite{lee2019meta} & ResNet-12 & 62.64$\pm$0.61& 78.63$\pm$0.46 & 65.99$\pm$0.72& 81.56$\pm$0.53\\ 
    Robust 20-distill$^{\flat\sharp\natural}$~\cite{dvornik2019diversity} & ResNet-18 & 63.06$\pm$0.61 & 80.63$\pm$0.42 & 65.43$\pm$0.21 & 70.44$\pm$0.32 \\ 
    SimpleShot$^{\flat}$~\cite{wang2019simpleshot} & ResNet-18 & 62.85$\pm$0.20 & 80.02$\pm$0.14 & 69.09$\pm$0.22 & 84.58$\pm$0.16\\ 
    CAN$^{\flat\heartsuit}$~\cite{hou2019cross} & ResNet-12 & 63.85$\pm$0.48 & 79.44$\pm$0.34 & 69.89$\pm$0.51 & 84.23$\pm$0.37\\ 
    
    S2M2$^{\flat\diamondsuit}$~\cite{mangla2020charting} & ResNet-18 & 64.06$\pm$0.18 & 80.58$\pm$0.12 & - & -\\ 
    Meta-Baseline$^{\flat}$~\cite{chen2020new}& ResNet-12 & 63.17$\pm$0.23 & 79.26$\pm$0.17 & 68.62$\pm$0.27 & 83.29$\pm$0.18 \\ 
    GNN+FT$^{\flat\heartsuit\natural}$~\cite{CrossDomainFewShot} & ResNet-10 & \textbf{66.32$\pm$0.80} & 81.98$\pm$0.55 & - & - \\ 
    DSN$^{\ddag}$~\cite{simon2020adaptive} & ResNet-12 & 62.64$\pm$0.66 & 78.83$\pm$0.45 & 66.22$\pm$0.75 & 82.79$\pm$0.48\\ 
    FEAT$^{\flat\heartsuit}$~\cite{ye2020fewshot} & ResNet-12 & \textbf{66.78$\pm$0.20}& {82.05$\pm$0.14} & {70.80$\pm$0.23}& \text{84.79$\pm$0.16}\\ 
    DeepEMD$^{\flat\diamondsuit\clubsuit}$~\cite{Zhang_2020_CVPR} & ResNet-12 & {65.91$\pm$0.82} & \textbf{82.41$\pm$0.56} & {71.16$\pm$0.87} & {86.03$\pm$0.58}\\ 
    Neg-Cosine$^{\flat\diamondsuit\natural}$ 	\cite{liu2020negative} & ResNet-12 & 63.85$\pm$0.81 &  81.57$\pm$0.56 & - & -\\ 
    Afrasiyabi $\etal^{\flat\heartsuit\diamondsuit\natural}$~\cite{Afrasiyabi_2020_ECCV} &ResNet-18 & 59.88$\pm$0.67 &80.35$\pm$0.73 & 69.29$\pm$0.56 &{85.97$\pm$0.49}\\ 
    E$^3$BM$^{\flat\heartsuit\diamondsuit}$~\cite{Liu2020E3BM} & ResNet-12 & 64.09$\pm$0.37 & 80.29$\pm$0.25 & 71.34$\pm$0.41 & 85.82$\pm$0.29 \\ 
    RFS-simple$^{\flat\ddag}$~\cite{tian2020rethinking} & ResNet-12 & 62.02$\pm$0.63 & 79.64$\pm$0.44 & 69.74$\pm$0.72 & 84.41$\pm$0.55\\ 
    RFS-distill$^{\flat\ddag\sharp}$~\cite{tian2020rethinking} & ResNet-12 & 64.82$\pm$0.60 & {82.14$\pm$0.43} & \textbf{71.52$\pm$0.69} & {86.03$\pm$0.49}\\ 
    \hline
    FRN (ours)$^{\flat}$ & ResNet-12 & {66.45$\pm$0.19} & \textbf{82.83$\pm$0.13} & {71.16$\pm$0.22} & {86.01$\pm$0.15}\\
    FRN (ours)$^{\flat\clubsuit}$ & ResNet-12 & - & - & \textbf{72.06$\pm$0.22} & \textbf{86.89$\pm$0.14} \\
    \hline
    \end{tabular}
    }
    \vspace{.8mm}
    \caption{Performance of selected competitive few-shot models on mini-ImageNet and tiered-ImageNet, ordered chronologically. 
    $\ddag$: use of data augmentation during evaluation; 
    $\sharp$: label smoothing, model ensemble or knowledge distillation; 
    $\heartsuit$: modules with many additional learnable parameters; 
    $\diamondsuit$: use of SGD during evaluation; 
    $\flat$: non-episodic pre-training or classifier losses; 
    $\natural$: network's input resolution is larger than 84. 
    $\clubsuit$: use of tiered-ImageNet data from DeepEMD's implementation. 
    } 
    \label{tab:imagenet}
    \end{table*}

We evaluate performance on two standard benchmarks. Compared to direct episodic meta-training from scratch, recent works~\cite{chen2020new,tian2020rethinking} gain a large advantage from pre-training on all training data and labels, followed by episodic fine-tuning. We follow the framework of~\cite{ye2020fewshot,Zhang_2020_CVPR} and pre-train our model on the entire training set as described in Sec.~\ref{sec:pretrain}.

\textbf{mini-ImageNet}~\cite{vinyals2016matching} is a subset of ImageNet containing 100 classes in total, with 600 examples per class. Following~\cite{ravi2016optimization}, we split categories into 64 classes for training, 16 for validation and 20 for test.

\textbf{tiered-ImageNet}~\cite{ren18fewshotssl} is a larger subset of ImageNet with 351-97-160 categories for training-validation-testing, respectively. 
Like tiered meta-iNat, tiered-ImageNet ensures larger domain differences between training and evaluation compared to mini-ImageNet. Most works~\cite{wang2019simpleshot,tian2020rethinking,chen2020new} use images from~\cite{ren18fewshotssl}\footnote{{\scriptsize\url{https://github.com/renmengye/few-shot-ssl-public}}} or~\cite{lee2019meta}\footnote{{\scriptsize\url{https://github.com/kjunelee/MetaOptNet}}}, which have 84$\times$84 resolution. DeepEMD~\cite{Zhang_2020_CVPR}'s implementation\footnote{{\scriptsize\url{https://github.com/icoz69/DeepEMD}}} has 224$\times$224 instead. For fair comparison, we experiment on both settings.

As shown in Table~\ref{tab:imagenet}, FRN is highly competitive with recent state-of-the-art results.
FRN leverages pre-training, but no other extra techniques or tricks. FRN also requires no gradient-based finetuning at inference time, which makes it more efficient than many existing baselines in practice.

\subsection{Cross-Domain Few-Shot Classification}
\begin{table}
\centering
\resizebox{.51\textwidth}{!}{
\hskip-.02\textwidth
\begin{tabular} {  l  c  c  c}
\hline
\textbf{Model} & \textbf{Backbone} & \textbf{1-shot} & \textbf{5-shot} \\
\hline
MAML$^{\natural\diamondsuit}$~\cite{finn2017model,chen2019closerfewshot} & ResNet-18 & - & 51.34$\pm$0.72 \\ 
ProtoNet$^{\natural}$~\cite{snell2017prototypical,chen2019closerfewshot} & ResNet-18 & - & 62.02$\pm$0.70 \\ 

Baseline$^{\flat\natural\diamondsuit}$~\cite{chen2019closerfewshot} & ResNet-18 & - & 65.57$\pm$0.70 \\ 
Baseline++$^{\flat\natural\diamondsuit}$~\cite{chen2019closerfewshot} & ResNet-18 & - & 62.04$\pm$0.76 \\ 
MetaOptNet$^{\sharp}$~\cite{lee2019meta,mangla2020charting} & ResNet-12 & 44.79$\pm$0.75 & 64.98$\pm$0.68\\ 
Diverse 20-full$^{\flat\sharp\natural}$~\cite{dvornik2019diversity} & ResNet-18 & - & 66.17$\pm$0.55 \\ 
SimpleShot$^{\flat}$~\cite{wang2019simpleshot,ziko2020laplacian} & ResNet-18 & 48.56 & 65.63 \\ 

MatchNet+FT$^{\flat\heartsuit\natural}$~\cite{CrossDomainFewShot} & ResNet-10 & 36.61$\pm$0.53 & 55.23$\pm$0.83\\ 
RelationNet+FT$^{\flat\heartsuit\natural}$~\cite{CrossDomainFewShot} & ResNet-10 & 44.07$\pm$0.77 & 59.46$\pm$0.71\\ 
GNN+FT$^{\flat\heartsuit\natural}$~\cite{CrossDomainFewShot} & ResNet-10 & 47.47$\pm$0.75 & 66.98$\pm$0.68\\ 
Neg-Softmax$^{\flat\diamondsuit\natural}$~\cite{liu2020negative} & ResNet-18 & - &69.30$\pm$0.73 \\ 
Afrasiyabi $\etal^{\flat\heartsuit\diamondsuit\natural}$~\cite{Afrasiyabi_2020_ECCV} &ResNet-18 &46.85$\pm$0.75& 70.37$\pm$1.02  \\ 
\hline
FRN (ours)$^{\flat}$ on classes from~\cite{chen2019closerfewshot} & ResNet-12 & \textbf{54.11$\pm$0.19} & \textbf{77.09$\pm$0.15} \\
FRN (ours)$^{\flat}$ on classes from~\cite{tang2020revisiting} & ResNet-12 & \textbf{51.60$\pm$0.21} & \textbf{72.97$\pm$0.18} \\
FRN (ours)$^{\flat}$ on all CUB classes & ResNet-12 & \textbf{53.39$\pm$0.21} & \textbf{75.16$\pm$0.17} \\ 
\hline
\end{tabular}
}
\vspace{.8mm}
\caption{Performance comparison in the cross-domain setting: mini-ImageNet$\rightarrow$CUB. 
Symbols and organization match Table~\ref{tab:imagenet}.
}
\label{tab:mini2cub}
\end{table}
Finally, we evaluate on the challenging cross-domain setting proposed by~\cite{chen2019closerfewshot}, where models trained on mini-ImageNet base classes are evaluated on test classes from CUB. 
We evaluate on three sets of CUB test classes: the split from~\cite{chen2019closerfewshot} (common in prior work), the split from~\cite{tang2020revisiting} (used in Sec.~\ref{sec:fine_grained}), and the full set of 200 CUB classes. 
As shown in Table~\ref{tab:mini2cub}, our FRN model from Sec.~\ref{sec:general} outperforms previous methods by a wide margin.

\section{Analysis}

\subsection{Ablation Study}
\label{sec:ablation}
    \textbf{Training shot}: Surprisingly, we found that FRN models trained on 1-shot episodes consistently underperform the same models trained with 5-shot episodes, even on 1-shot evaluation. We therefore report the superior numbers from 5-shot models in Sec.~\ref{sec:experiments} and include the 1-shot performance as an ablation in Table~\ref{tab:shots} of SM. 
    Though clearly worse than the 5-shot counterpart, 1-shot FRN is broadly competitive with the best-performing baselines.

    \textbf{Pre-training}: FRN pre-training is crucial for competitive general few-shot performance, especially when compared to pre-trained baselines. However, pre-training alone does not produce a competitive few-shot learner. An FRN trained from scratch outperforms a pre-trained FRN evaluated naively (Table~\ref{tab:ablation_pretrain}, bottom rows). The two-round process of pre-training followed by episodic fine-tuning appears to be crucial. This finding is in line with prior work~\cite{ye2020fewshot,chen2020new}. 
    
    Classical pre-training with average-pooling, however, does not produce a viable classifier (Table~\ref{tab:ablation_pretrain}, middle rows). 
    These features are not spatially distinct enough for FRN fine-tuning to recover a meaningful feature space. The resulting classifier is \textit{worse} than one trained from scratch.

\begin{table}
\centering
\resizebox{.41\textwidth}{!}{
\scriptsize
\setlength\tabcolsep{5pt}
\hskip-.02\textwidth
\begin{tabular} {  l  c  c}
\hline
\textbf{training setting} & \textbf{1-shot} & \textbf{5-shot} \\
\hline
episodic train from scratch & 63.03$\pm$0.20  & 78.01$\pm$0.15  \\
\hline
after avg-pool pre-train & 59.43$\pm$0.20 & 70.88$\pm$0.16 \\
episodic finetune & 62.13$\pm$0.20  & 76.28$\pm$0.15 \\
\hline
after FRN pre-train & 60.97$\pm$0.21 & 75.11$\pm$0.18 \\
episodic finetune & \textbf{66.45$\pm$0.19}  & \textbf{82.83$\pm$0.13} \\
\hline
\end{tabular}
}
\vspace{.8mm}
\caption{Impact of pre-training for FRN on mini-ImageNet. Both pre-training and episodic finetuning are important. Classical pre-training with global average-pooling works poorly.}
\label{tab:ablation_pretrain}
\end{table}

    \textbf{Auxiliary loss and $\lambda,\rho$ regularizers}: We ablate these components of both Conv-4 and ResNet-12 models on cropped CUB, with results in Table~\ref{tab:ablation_cub} in SM. 
    The auxiliary loss has little to no impact on FRN performance -- we include this loss in our experimental models only for consistent comparisons. Fixing $\alpha,\beta$ to 0 (and thus $\lambda,\rho$ to constants) yields mixed results. The 4-layer network clearly benefits from learning both values, but the ResNet-12 architecture does not, likely because the high-dimensional feature space is rich enough to overcome any regularization problems on its own.

\subsection{Reconstruction Visualization}
\label{sec:viz}

    While our results suggest that FRN produces more semantically faithful reconstructions from same-class support images than from different classes, we would like to confirm this visually. 
    We therefore train image re-generators for the 5-shot ResNet-12 FRN on CUB and mini-ImageNet, which use an inverted ResNet-12 to map FRN features back to the original image. Training details can be found in SM Sec.~\ref{sup:decoder}. Results are reported on validation images.
    
    \begin{table}
    \centering
    \resizebox{.35\textwidth}{!}{
    \scriptsize
    \setlength\tabcolsep{5pt}
    \hskip-.02\textwidth
    \begin{tabular} {  l  c  c}
    \hline
    \textbf{Input} & \textbf{CUB} & \textbf{mini-IN}\\
    \hline
    ground-truth feature map & .208 & .177 \\
    same-class reconstruction & .343 & .307 \\
    diff-class reconstruction & .385 & .337 \\
    \hline
    \end{tabular}
    }
    \vspace{.8mm}
    \caption{L2 pixel error between original images and regenerated images from different latent inputs.
    Results are averaged over 1,000 trials and 95\% confidence intervals are below 1e-3.}
    \label{tab:regenerate}
    \end{table}
    
\begin{figure}
    \centering
    \includegraphics[width=\linewidth]{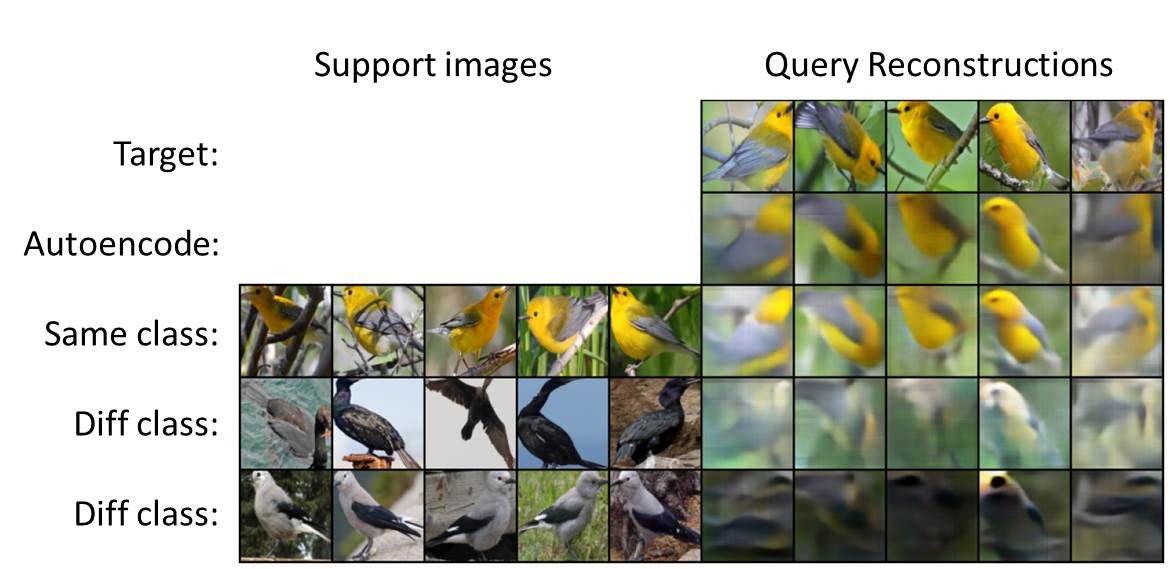}
    \caption{CUB images are regenerated from ground-truth feature maps (row 2), and reconstructions from same-class (row 3) and different-class support images (rows 4, 5). Same-class reconstructions are more faithful to the original. Best viewed digitally.}
    \label{fig:regenerate}
\end{figure}

    If same-class feature map reconstructions are more semantically faithful than different-class ones, we should observe a corresponding difference in regenerated image quality. 
    Fig.~\ref{fig:regenerate} and Table~\ref{tab:regenerate} verify this. 
    Reconstructions from ground-truth features are not particularly good, as classifiers discard class-irrelevant details. 
    However, the increase in ground-truth pixel error relative to these target feature maps is clearly smaller for same-class reconstructions.  
    Additional visualizations are provided in SM Sec.~\ref{sup:visual}. 
    We conclude that FRN reconstructions are semantically faithful for same-class support images and less faithful otherwise.  

\section{Conclusion}
We introduce Feature Map Reconstruction Networks, a novel approach to few-shot classification based on reconstructing query features in latent space. Solving the reconstruction problem in closed form produces a classifier that is both straightforward and powerful, incorporating fine spatial details without overfitting to position or pose. We demonstrate state-of-the-art performance on four fine-grained few-shot classification benchmarks, and highly competitive performance in the general setting. 

\paragraph{Acknowledgements}: This work was funded by the DARPA Learning with Less Labels program (HR001118S0044). The authors would like to thank Cheng Perng Phoo for valuable suggestions and assistance.

{\small
\bibliographystyle{ieee_fullname}
\bibliography{egbib}
}

\onecolumn

\begin{center}
    \LARGE
    \textbf{Supplementary Materials}
\end{center}

\section{Pseudo-Codes for Calculating $\bar Q_c$}
\label{sup:code}

Eq.~\ref{eq:q_alter}: $\bar Q_c = \rho \bar WS_c = \rho Q(S_c^TS_c+\lambda I)^{-1}S_c^TS_c$. It is more computation-efficient than Eq.~\ref{eq:q_full} when $kr>d$. For consistency, we use this formula in our implementation for all experiments. The corresponding Pytorch-style pseudo-code can be found in Listing~\ref{code_10}.

\begin{lstlisting}[language=Python, caption={Pytorch pseudo-code for feature map reconstruction in Eq.~\ref{eq:q_alter} for a single meta-training episode. The whole calculation can be performed in parallel via batched matrix multiplication and inversion.},label={code_10}]
# input tensors and scalars:
# query: feature maps of query images, the tensor shape is [way*query_shot*r, d] 
# support: feature maps of support images, the tensor shape is [way, support_shot*r, d]
# lam: lambda
# rho

# output: Q_bar


def get_Q_bar(query,support,lam,rho):

    st = support.permute(0,2,1)
    sts = st.matmul(support)
    m_inv = (sts+torch.eye(sts.size(-1)).unsqueeze(0).mul(lam)).inverse() 
    hat = m_inv.matmul(sts)
    Q_bar = query.matmul(hat).mul(rho)
    
    return Q_bar
\end{lstlisting}

Eq.~\ref{eq:q_full}: $\bar Q_c = \rho \bar W S_c = \rho QS_c^T(S_cS_c^T+\lambda I)^{-1}S_c$. It is more computation-efficient than Eq.~\ref{eq:q_alter} when $d>kr$. The corresponding Pytorch-style pseudo-code can be found in Listing~\ref{code_8}.

\begin{lstlisting}[language=Python, caption={Pytorch pseudo-code for feature map reconstruction in Eq.~\ref{eq:q_full} for a single meta-training episode. The whole calculation can be performed in parallel via batched matrix multiplication and inversion. Details about inputs can be found in Listing~\ref{code_8}.},label={code_8}]
def get_Q_bar(query,support,lam,rho):
    
    st = support.permute(0,2,1)
    sst = support.matmul(st)
    m_inv = (sst+torch.eye(sst.size(-1)).unsqueeze(0).mul(lam)).inverse()
    hat = st.matmul(m_inv).matmul(support)
    proj = query.matmul(hat).mul(rho)
    
    return Q_bar
\end{lstlisting}

\section{Ablation Studies}
\label{sup:ablation}
Numerical results and deeper discussion from the main paper.

\subsection{Episodic Training Shot Number}
\label{sup:shots}
In our experiments, we surprisingly found that FRN trained with 5-shot episodes consistently outperforms FRN trained with 1-shot episodes, even on 1-shot evaluation. 
For the sake of brevity, we only report the superior performance from the FRN trained with 5-shot episodes in the main paper and include the performance of FRN trained with 1-shot episodes as an ablation in Table~\ref{tab:shots}. 
We also include the best-performing baseline from each experimental setting in the main paper for comparison. 
Under the standard 5-way 1-shot evaluation, 1-shot trained FRN outperforms the baseline in most fine-grained settings, but also underperforms the 5-shot trained models in all settings.

\begin{table}[h]
\centering
\resizebox{.97\textwidth}{!}{
\scriptsize
\setlength\tabcolsep{5pt}
\hskip-.02\textwidth
\begin{tabular} {  l  c  c  c  c  c c } 
\hline
\textbf{Model (Conv-4)} & \textbf{Training scheme}& \multicolumn{2}{c}{\textbf{CUB (crop)}} & \textbf{Aircraft} & \textbf{meta-iNat} & \textbf{tiered meta-iNat}\\
\hline
Best Baseline & 1-shot & \multicolumn{2}{c}{69.64$\pm$0.23} & 49.67$\pm$0.22 & 60.03$\pm$0.23 & 36.83$\pm$0.18 \\
FRN (ours) & 1-shot & \multicolumn{2}{c}{68.27$\pm$0.23} & 50.25$\pm$0.21  & 60.04$\pm$0.22 & 40.11$\pm$0.18 \\
FRN (ours) & 5-shot & \multicolumn{2}{c}{73.48$\pm$0.21} & 53.20$\pm$0.21  & 62.42$\pm$0.22 & 43.91$\pm$0.19 \\
\hline

\hline
\textbf{Model (ResNet-12)} & \textbf{Training scheme} & \textbf{CUB (crop)} & \textbf{CUB (uncrop)} & \textbf{Aircraft} & \textbf{mini-ImageNet} & \textbf{tiered-ImageNet}\\
\hline
Best Baseline & 1-shot & 80.80$\pm$0.20 & 79.96$\pm$0.21 & 68.16$\pm$0.23 & 66.78$\pm$0.20 & 71.52$\pm$0.69 \\
FRN (ours) & 1-shot & 82.62$\pm$0.19 & 82.02$\pm$0.20 & 69.40$\pm$0.23  & 65.33$\pm$0.20 & 70.88$\pm$0.22 \\
FRN (ours) & 5-shot & 83.16$\pm$0.19 & 83.55$\pm$0.19 & 70.17$\pm$0.22  & 66.45$\pm$0.19 & 72.06$\pm$0.22 \\
\hline
\end{tabular}
}
\vspace{.8mm}
\caption{5-way 1-shot performance comparison for FRN trained with 1-shot and 5-shot episodes, along with the best reported baseline for reference. 
Results for FRN are averaged over 10,000 trials with 95\% confidence interval. 
}
\label{tab:shots}
\end{table}

\subsection{Auxiliary Loss and Regularization Parameters}
    Ablation results for these components on cropped CUB are given in Table~\ref{tab:ablation_cub}. 
    In this case, 1-shot results come from models trained with 1-shot episodes (as in Sec.~\ref{sup:shots}), rather than 5-shot (as in the main paper). 
    We find that the auxiliary loss has little to no consistent impact on FRN performance. We include it for our experimental models only for the sake of apples-to-apples comparison with baseline models that do rely on it, such as DSN~\cite{simon2020adaptive} (we verified in early experiments that our proxy DSN$^{\dag}$ implementation also suffers when the auxiliary loss is removed). 
    
    To analyze the contribution of the learned regularization terms $\lambda$ and $\rho$, we disable their learnability, setting $\alpha$, $\beta$, or both equal to zero over the course of training. 
    The impact of these terms is mixed. The 4-layer network clearly benefits from learning these terms, but the ResNet-12 architecture benefits from removing them -- though by only a small, possibly insignificant margin. It seems that the more powerful network is already able to overcome any regularization issues, by massaging the feature space in a more elegant way than the individual $\lambda,\rho$ terms can provide.

\begin{table}
\centering
\resizebox{.6\textwidth}{!}{
\scriptsize
\hskip-.01\textwidth
\begin{tabular} {  l  c c c c}
\hline
\-\ & \multicolumn{2}{c}{\textbf{Conv-4}} & \multicolumn{2}{c}{\textbf{ResNet-12}} \\
\textbf{Model} & \textbf{1-shot} & \textbf{5-shot} & \textbf{1-shot} & \textbf{5-shot}\\
\hline
no Aux& \textbf{68.38$\pm$0.23} &	88.13$\pm$0.13 & 82.20$\pm$0.20 &	92.65$\pm$0.10\\
fixed $\lambda$ & 66.44$\pm$0.23	& 84.05$\pm$0.15 & 82.40$\pm$0.20 & 93.12$\pm$0.10\\
fixed $\rho$ & 67.34$\pm$0.23	& 87.60$\pm$0.13 & 82.58$\pm$0.20	& 92.79$\pm$0.10\\
fixed $\lambda$, $\rho$ & 66.19$\pm$0.23 & 83.27$\pm$0.15	& \textbf{82.80$\pm$0.19} & \textbf{93.33$\pm$0.10}\\
\hline
whole model& 68.27$\pm$0.23 & \textbf{88.43$\pm$0.13} & 82.62$\pm$0.19 & 92.59$\pm$0.10 \\
\hline
\end{tabular}
}
\vspace{.8mm}
\caption{Ablation study on FRN regularization parameters and auxiliary loss using the cropped CUB benchmark.}
\label{tab:ablation_cub}
\end{table}

\section{Training Details}

\label{sup:training}
    \textbf{Pre-processing:} During training on Aircraft, CUB, mini-ImageNet and tiered-ImageNet (data from DeepEMD~\cite{Zhang_2020_CVPR}'s implementation\footnote{{\scriptsize\url{https://github.com/icoz69/DeepEMD}}}), images are randomly cropped and then resized into 84$\times$84\footnote{i.e., \texttt{torchvision.transforms.RandomResizedCrop(84)}}. For meta-iNat, tiered meta-iNat, and tiered-ImageNet using the original release by Ren~\etal~\cite{ren18fewshotssl}\footnote{\scriptsize\url{https://github.com/renmengye/few-shot-ssl-public}}, images are randomly padded then cropped into 84$\times$84\footnote{i.e., \texttt{torchvision.transforms.RandomCrop(84, padding=8)}}. Then all images are augmented by color-jittering and random horizontal flipping.
    
    During inference on Aircraft and CUB (using bounding-box cropped images as input), images are resized into 92$\times$92 then center-cropped into 84$\times$84\footnote{i.e., \texttt{torchvision.transforms.Resize([92,92])} followed by \texttt{torchvision.transforms.CenterCrop(84)}}. For mini-ImageNet, tiered-ImageNet (data from DeepEMD~\cite{Zhang_2020_CVPR}'s implementation) and CUB (using raw images as input), images are resized such that the smaller edges will match the length of 92\footnote{i.e., \texttt{torchvision.transforms.Resize(92)}} then center-cropped into 84$\times$84. This prevents distortion artifacts at test time. For meta-iNat, tiered meta-iNat, and tiered-ImageNet using the original release by Ren~\etal~\cite{ren18fewshotssl}, images are already at size 84$\times$84 and are therefore fed into models directly. 
    
    Many of these choices were inherited from the original datasets and code. They do not reflect hyperparameter tuning on any of our models. 
    
    \textbf{Network Backbones:} We use two neural architectures in our experiments: Conv-4 and ResNet-12. Conv-4 consists of four convolutional layers with kernel size $3\times3$ and output size 64, followed by BatchNorm, ReLU, and $2\times 2$ max-pooling. ResNet-12 consists of four residual blocks, each with three convolutional layers, with leaky ReLU (0.1) and $2\times 2$ max-pooling on the main stem. We use drop-block as in the original implementation~\cite{tian2020rethinking,ye2020fewshot,lee2019meta}\footnote{{\scriptsize\url{https://github.com/WangYueFt/rfs}}}\footnote{{\scriptsize\url{https://github.com/Sha-Lab/FEAT}}}\footnote{{\scriptsize\url{https://github.com/kjunelee/MetaOptNet}}}.
    Output sizes for each residual block are: $64, 160, 320, 640$. 
    
    \textbf{Hyperparameters:} Unless otherwise stated, all models are trained with a weight decay of 5e-4. 
    We temperature scale all output probability logits, and also normalize these logits for our baseline models, dividing by $64$ for Conv-4 and $640$ for ResNet-12.
    FRN models do not benefit from this normalization, as logit scale does not correspond well to embedding scale due to the large influence of the regularization term. 
    We found it necessary for stable training to instead normalize the FRN ResNet-12 \textit{embeddings} by downscaling by a factor of $\sqrt{640}$. 
    Note that this is algebraically equivalent to the existing logit-based normalization for the ProtoNet$^{\dag}$ and DSN$^{\dag}$ baselines, as well as CTX$^{\dag}$ up to a rescaling factor in the inner softmax term (see Eq.~\ref{eq:ctx}).  
    
    Training details for each individual benchmark are provided below. 
    We found that our prototypical network hyperparameters worked well for all baselines, and for fair comparison, do not tune FRN hyperparameters separately (beyond addressing obvious cases of training instability). Thus we give one set of hyperparameters for all models, with exceptions noted. 
    
    While we provide training details for 1-shot models, note that in our benchmark experiments these are only relevant to our implemented baselines (i.e. ProtoNet$^{\dag}$, DSN$^{\dag}$ and CTX$^{\dag}$). 1-shot trained FRN models appear only in the ablation studies of Sec.~\ref{sec:ablation} and Sec.~\ref{sup:ablation}. FRN results reported in the main paper all come from models trained with 5-shot episodes.

\subsection{Fine-Grained Few-Shot Benchmarks}
\label{sup:fine-grained}
    \textbf{CUB:} Our implementation follows~\cite{gidaris2018dynamic,lee2019meta}. We train all Conv-4 models for 800 epochs using SGD with Nesterov momentum of 0.9 and an initial learning rate of 0.1. The learning rate decreases by a factor of 10 at epoch 400. ResNet-12 models train for 1200 epochs and scale down the learning rate at epochs 400 and 800. We use the validation set to select the best-performing model over the course of training, and validate every 20 epochs. 
    
    Conv-4 models are trained with the standard episodic setup: 20-way 5-shot for 5-shot models, and 30-way 1-shot for 1-shot models. We use 15 query images per class in both settings. In order to save memory we cut these values in half for ResNet-12 models: 5-shot models train on 10-way episodes, while 1-shot models train on 15-way episodes.

    \textbf{Aircraft:} Our implementation for the aircraft dataset roughly follows the CUB setup. Conv-4 models train for 1200 epochs, however, and we found that the 1-shot ResNet-12 ablation FRN is highly unstable at the very beginning of training and frequently collapses to the suboptimal uniform solution. We found it necessary to train this model using Adam with an initial learning rate of 1e-3, and no weight decay. All other hyperparameters remain the same as for CUB.

    \textbf{meta-iNat and tiered meta-iNat:} Our implementation follows~\cite{wertheimer2019few}, only we train for twice as long, as we found that the loss curves frequently failed to stabilize before each decrease in learning rate. We therefore train our models for 100 epochs using Adam with initial learning rate 1e-3. We cut the learning rate by a factor of two every 20 episodes. Because there is no validation set available, we simply use the final model at the end of training. Episode setup is as in CUB and Aircraft. 

\subsection{General Few-Shot Benchmarks}
\label{app:general}
    \textbf{mini-ImageNet:} For non-episodic FRN pre-training, we run 350 epochs with batch size 128, using SGD with initial learning rate 0.1 and Nesterov momentum 0.9. We cut the learning rate by a factor of 10 at epochs 200 and 300. We also use the validation set to select the best-performing model over the course of pre-training, and validate every 25 epochs. Subsequent episodic fine-tuning uses the same optimizer for 150 epochs, but with initial learning rate 1e-3, decreased by a factor of 10 at epochs 70 and 120. For the 5-shot FRN model, we use the standard 20-way episodes. The 1-shot ablation model uses 25-way episodes. 

    FRN models trained from scratch (for the ablation study) use the same optimizer for 300 epochs, with initial learning rate 0.1, decreased by a factor of ten at epochs 160 and 250. Similar to CUB, we cut the way of the training episodes in half in order to reduce memory footprint: 10-way for 5-shot models, and 15-way for 1-shot ablation models. 
    
    For all the episodic meta-training processes mentioned above, we continue to use the validation set to select the best-performing model by validating every 5 epochs for fine-tuning and every 20 epochs for training from scratch. 
    
    \textbf{tiered-ImageNet:} For non-episodic FRN pre-training, we run 90 epochs with batch size 128, using SGD with initial learning rate 0.1 and Nesterov momentum 0.9. We cut the learning rate by a factor of 10 at epochs 30 and 60. We also use the validation set to select the best-performing model over the course of pre-training, and validate every 5 epochs. Subsequent episodic fine-tuning uses the same optimizer for 60 epochs, but with initial learning rate 1e-3, decreased by a factor of 10 at epochs 30 and 50. For the 5-shot FRN model, we use the standard 20-way episodes. The 1-shot ablation model uses 25-way episodes. We also use the validation set to select the best-performing model during episodic fine-tuning, validating every 5 epochs. 
    
    For the experiment on the tiered-ImageNet dataset provided by DeepEMD~\cite{Zhang_2020_CVPR}'s implementation\footnote{{\scriptsize\url{https://github.com/icoz69/DeepEMD}}}, since images have larger resolution, we train slightly longer during the episodic fine-tuning: 70 epochs in total, learning rate decreased by a factor of 10 at epochs 40 and 60. All other hyperparameters remain the same as above.

\subsection{Implemented Baselines}
\label{sup:comparison}
    While our implemented baselines closely mirror the models that appear in prior work, we do introduce slight differences in order to provide the most direct comparison to FRN as possible. Generally, we found these modifications actually improved performance. We enumerate differences baseline by baseline. 
    
    \textbf{ProtoNet$^{\dag}$}:
    All of our baselines, including prototypical networks, incorporate temperature scaling with initial value set to $\frac{1}{d}$, where $d$ is the latent dimensionality of the feature extractor. While the original prototypical network paper~\cite{snell2017prototypical} did not incorporate this practice, it has become common in subsequent work and has been shown to improve performance~\cite{gidaris2018dynamic,lee2019meta,ye2020fewshot}. 
    
    \textbf{DSN$^{\dag}$}:
    Our implementation of DSN is mathematically equivalent to the original~\cite{simon2020adaptive}, but is implemented differently. Our method calculates the closed-form solution to the regression objective, which is equivalent to calculating the projection of the query point into the subspace defined by the supports (for sufficiently small regularizer $\lambda$; we use 0.01). \cite{simon2020adaptive} instead calculates an orthonormal basis that spans the support points using SVD, and zeros out the components of the query point that fall into that subspace, yielding the displacement from the projection. While the underlying mechanics are different, the end result is the same. 
    
    The original DSN also recenters the projection problem about each class centroid -- that is, DSN projects each query point onto the hyperplane containing the support points but not necessarily the origin. In our implementation and in FRN, we define the projection subspace as the hyperplane containing the support points and the origin (that is, we maintain the origin as a common reference point). We found this choice made little difference in practice, and including the origin / not recentering actually improved DSN performance slightly.  
    
    Finally, the true DSN auxiliary loss encourages orthogonality between the calculated orthonormal bases for each class. This is a cheap operation for the original DSN model, as these bases are pre-calculated from the prediction step. This is not true for our model; we instead utilize the same loss formula in Eq.~\ref{eq:aux} but compare the actual (normalized) support points rather than their orthonormal bases. 
    
    \textbf{CTX$^{\dag}$}: 
    For fair comparison, we do not include SimCLR training episodes as in the original implementation \cite{doersch2020crosstransformers}, as this can also be employed by our other baselines and by FRN itself. Instead, we re-use the auxiliary loss adapted from DSN.

\subsection{Image Decoder}
\label{sup:decoder}
    The decoder network for the visualizations in Section~\ref{sec:viz} takes the form of an inverted ResNet-12, with four residual blocks and a final projection layer to three channels. Upsampling is performed using strided transposed convolution in both the residual stem and the main stem of each residual block. These transposed convolution layers also downsize the number of channels. The final residual block outputs of size 80$\times$80 are rescaled to the original 84$\times$84 resolution using bilinear sampling before the final projection layer. The full architecture is described in Table~\ref{tab:vizform}. 
    
    \begin{table}
    \centering
    \small
    \setlength\tabcolsep{5pt}
    \hskip-.02\textwidth
    \begin{tabular} {| c | c |}
    \hline
    \multicolumn{2}{|c|}{\textbf{Decoder Network:}}\\
    \hline
    \textbf{Layer} & \textbf{Output Size}\\
    \hline
    Input & $640\times5\times5$\\
    \hline
    ResBlock & $512\times10\times10$\\
    \hline
    ResBlock & $256\times20\times20$\\
    \hline
    ResBlock & $128\times40\times40$\\
    \hline
    ResBlock & $64\times80\times80$\\
    \hline
    Upsample & $64\times84\times84$\\
    \hline
    Conv3x3 & $3\times84\times84$\\
    \hline
    Tanh & $3\times84\times84$\\
    \hline
    \end{tabular}
    \hspace{2em}
    \begin{tabular} {| c | c |}
    \hline
    \multicolumn{2}{|c|}{\textbf{Residual Block:}}\\
    \hline
    \multicolumn{2}{|c|}{Input}\\
    \hline
    \multirow{4}{*}{TranspConv4x4, stride 2} & Conv3x3\\
    \cline{2-2}
    & BatchNorm\\
    \cline{2-2}
    & ReLU\\
    \cline{2-2}
    & TranspConv4x4, stride 2\\
    \hline
    \multirow{4}{*}{BatchNorm} & BatchNorm\\
    \cline{2-2}
    & ReLU\\
    \cline{2-2}
    & Conv3x3\\
    \cline{2-2}
    & BatchNorm\\
    \hline
    \multicolumn{2}{|c|}{Addition}\\
    \hline
    \multicolumn{2}{|c|}{ELU}\\
    \hline
    \end{tabular}
    \vspace{.8mm}
    \caption{Neural architecture for the image decoder network (left) and residual block (right). Transposed convolution layers have half as many output channels as input channels.}
    \label{tab:vizform}
    \end{table}
    
    The decoder network is trained through L-1 reconstruction loss using Adam with an initial learning rate of 0.01 and batch size of 200. The CUB decoder is trained for 500 epochs, with learning rate decreasing by a factor of 4 every 100 epochs. mini-ImageNet is a larger dataset than CUB, so the corresponding decoder is trained for 200 epochs, with learning rate decreasing every 40 epochs. 
    
    Over the course of training, we found that the FRN classifier learns to regularize the reconstruction problem heavily. This is problematic in that the regularized reconstructions all fall off the input manifold for the decoder network, producing uniformly flat grey-brown images. To prevent this, we removed the normalization on the classifier latent embeddings (multiplying all inputs by $\sqrt{640}$). This was sufficient to eliminate the negative impact of regularization and produce meaningful image reconstructions. 
\section{Computational Efficiency}
\label{sup:speed}

\textbf{Speed comparison to DeepEMD:} As shown in Table~\ref{tab:speed}, we compare the computation speed between FRN and DeepEMD~\cite{Zhang_2020_CVPR} with different episode shot numbers and implementation variants. Both use ResNet-12 as backbones and are evaluated on mini-ImageNet. For each episode, each class contains 15 query images during meta-training, and 16 query images during meta-testing. FRN is much more efficient in both training and evaluation scenarios.

For DeepEMD, the OpenCV~\cite{opencv_library} implementation of the EMD solver uses a modified Simplex algorithm to calculate the transport matrix, while qpth~\cite{amos2017optnet} uses an interior point method. It should be noted that while the OpenCV solver is faster, its gradients cannot be accessed. Thus it should not be used for meta-training (while it is possible, the calculated transport matrix must be treated like a constant and so gradient quality -- and downstream performance -- suffers). 
Note that for meta-testing episodes with shot number above one, DeepEMD learns a structured fully-connected layer via SGD. This leads to a significant slowdown compared to the 1-shot case -- both because multiple forward and backward passes are required, and because once again OpenCV cannot provide gradients and thus should not be used. 

For the sake of completeness, we provide speed results for both options of FRN's implementation (Eq.~\ref{eq:q_full} and Eq.~\ref{eq:q_alter}). For ResNet-12, $d=640$, $k=1$ or $5$, and $r=5\times 5=25$, thus $d>kr$. From the analysis in Sec.~\ref{sec:alter}, Eq.~\ref{eq:q_full} should then be more efficient than Eq.~\ref{eq:q_alter}, and this can be verified in Table~\ref{tab:speed}. All other experiments use Eq.~\ref{eq:q_alter} as FRN's implementation, however, only for the sake of consistency.

\textbf{Memory comparisons to CTX$^{\dag}$:} 
As shown in Table~\ref{tab:memory}, we compare the GPU memory usage between FRN and CTX$^{\dag}$ during training on CUB, following the setup described in Sec.~\ref{sup:fine-grained}. We use Conv-4 models and upscale the feature map resolution to 10$\times$10 to investigate how performance changes as $kr$ increases. Memory is reported on a single GPU using the \texttt{nvidia-smi} resource tracker. While differences in the 5$\times$5 setting are mostly insignificant, the 10$\times$10 setting shows a rapid blowup in memory usage for CTX$^{\dag}$ that FRN is able to mitigate. We include both formulations of FRN for completeness.
Interestingly, the analysis in Sec.~\ref{sec:alter} breaks down in this setting: contrary to expectations, 1-shot memory usage is the same in both FRN variants, and Eq.~\ref{eq:q_full} is more efficient than Eq.~\ref{eq:q_alter} in the 5-shot $5\times 5$ setting, despite the fact that it is inverting a larger matrix ($125\times 125$ vs $64\times 64$). 
This could be due to particular implementations of matrix inversion, multiplication, and axis permutation on the GPU. 
However, we need not choose one variant over the other \textit{a priori}. As both variants are algebraically equivalent, they can be substituted on the fly.

\begin{table}
\centering
\resizebox{.7\textwidth}{!}{
\scriptsize
\setlength\tabcolsep{5pt}
\hskip-.02\textwidth
\begin{tabular} {  l  c  c  c  c}
\hline
\textbf{Model} & \textbf{Implementation}& \textbf{Phase} & \textbf{Episode} & \textbf{Time per Episode (ms)} \\
\hline
DeepEMD~\cite{Zhang_2020_CVPR} & qpth~\cite{amos2017optnet} & meta-train & 5-way 1-shot & 26,000  \\
FRN (ours) & Eq.~\ref{eq:q_alter} & meta-train & 5-way 1-shot & \textbf{281} \\
FRN (ours) & Eq.~\ref{eq:q_full} & meta-train & 5-way 1-shot & \textbf{279} \\
\hline
DeepEMD~\cite{Zhang_2020_CVPR} & qpth~\cite{amos2017optnet} & meta-train & 5-way 5-shot & $>$990,000  \\
FRN (ours) & Eq.~\ref{eq:q_alter} & meta-train & 5-way 5-shot & \textbf{357} \\
FRN (ours) & Eq.~\ref{eq:q_full} & meta-train & 5-way 5-shot & \textbf{346} \\
\hline
DeepEMD~\cite{Zhang_2020_CVPR} & qpth~\cite{amos2017optnet} & meta-test & 5-way 1-shot & 23,275  \\
DeepEMD~\cite{Zhang_2020_CVPR} & OpenCV~\cite{opencv_library} & meta-test & 5-way 1-shot & 178  \\
FRN (ours) & Eq.~\ref{eq:q_alter} & meta-test & 5-way 1-shot & \textbf{73} \\
FRN (ours) & Eq.~\ref{eq:q_full} & meta-test & 5-way 1-shot & \textbf{63} \\
\hline
DeepEMD~\cite{Zhang_2020_CVPR} & qpth~\cite{amos2017optnet} & meta-test & 5-way 5-shot & $>$800,000  \\
DeepEMD~\cite{Zhang_2020_CVPR} & OpenCV~\cite{opencv_library} & meta-test & 5-way 5-shot & 18,292  \\
FRN (ours) & Eq.~\ref{eq:q_alter} & meta-test & 5-way 5-shot & \textbf{88} \\
FRN (ours) & Eq.~\ref{eq:q_full} & meta-test & 5-way 5-shot & \textbf{79} \\
\hline
\end{tabular}
}
\vspace{.8mm}
\caption{Speed comparison between DeepEMD and FRN on mini-ImageNet with ResNet-12 as the backbone. FRN outperforms DeepEMD with a large margin for different shot numbers under both meta-training and meta-testing phases.
}
\label{tab:speed}
\end{table}

\begin{table}
\centering
\resizebox{.4\textwidth}{!}{
\scriptsize
\setlength\tabcolsep{5pt}
\hskip-.02\textwidth
\begin{tabular} {  l  c  c  c  c}
\hline
\-\ & \multicolumn{2}{c}{\textbf{5$\times$5}} & \multicolumn{2}{c}{\textbf{10$\times$10}}\\
\textbf{Model} & {1 shot}& {5 shot} & {1 shot} & {5 shot} \\
\hline
CTX$^{\dag}$ & 6225 & 5449 & 8979 & 9693\\
FRN (Eq.~\ref{eq:q_full}) & 6305 & 5317 & 7125 & 6767 \\
FRN (Eq.~\ref{eq:q_alter}) & 6305 & 6529 & 7125 & 6727\\
\hline
\end{tabular}
}
\vspace{.8mm}
\caption{Memory usage in megabytes for training CTX$^{\dag}$ and FRN. While differences are small in the 5$\times$5 feature map resolution setting, increasing the resolution to 10$\times$10 is sufficient to produce large differences in memory usage.
Note that these numbers reflect convolutional network layers as well as the final reconstruction layer.
}
\label{tab:memory}
\end{table}

\section{Additional Results on CUB}
\label{sup:cub}

\subsection{Class Split}
Unlike mini-ImageNet, for which researchers tend to use the same class split as~\cite{ravi2016optimization}, CUB doesn't have an official split. Although practitioners agree on the train/validation/test split ratio (i.e., 100/50/50), there exist many different specific class split implementations~\cite{reed2016learning, triantafillou2017few, chen2019closerfewshot, ye2020fewshot, tang2020revisiting}. 
For our CUB experiments in the main paper, we use the same random train/validation/test split as Tang~\etal~\cite{tang2020revisiting}. 
Many recent open-sourced works~\cite{ziko2020laplacian,liu2020negative,mangla2020charting} use the dataset-generating code\footnote{{\scriptsize\url{https://github.com/wyharveychen/CloserLookFewShot/blob/master/filelists/CUB/write_CUB_filelist.py}}} from Chen~\etal~\cite{chen2019closerfewshot}, using raw, non-cropped images as input. Therefore, we re-run our method on CUB under the uncropped setting using the same class split as Chen~\etal~\cite{chen2019closerfewshot}. As shown in Table~\ref{tab:closer_split}, FRN performance is not impacted by the choice of class split.

\begin{table}
\centering
\resizebox{.65\textwidth}{!}{
\scriptsize
\setlength\tabcolsep{5pt}
\hskip-.02\textwidth
\begin{tabular} {  l  c  c  c  c}
\hline
\textbf{Model} & \textbf{class split}& \textbf{Backbone} & \textbf{1-shot} & \textbf{5-shot} \\
\hline
FRN (ours) & Tang~\etal~\cite{tang2020revisiting}& ResNet-12& \textbf{83.55$\pm$0.19}  & \textbf{92.92$\pm$0.10}  \\
FRN (ours) &Chen~\etal~\cite{chen2019closerfewshot}& ResNet-12& \textbf{83.35$\pm$0.18}  & \textbf{93.28$\pm$0.09} \\
\hline
        Baseline++$^{\flat}$~\cite{chen2019closerfewshot} &Chen~\etal~~\cite{chen2019closerfewshot}& ResNet-34 & 68.00$\pm$0.83 & 84.50$\pm$0.51  \\
        ProtoNet~\cite{chen2019closerfewshot,snell2017prototypical} &Chen~\etal~\cite{chen2019closerfewshot}&	ResNet-34&	72.94$\pm$0.91&	87.86$\pm$0.47\\
        S2M2$^{\flat}$~\cite{mangla2020charting} &Chen~\etal~\cite{chen2019closerfewshot}& WRN-28-10 & 80.68$\pm$0.81 & 90.85$\pm$0.44\\
        Neg-Cosine$^{\flat}$~\cite{liu2020negative} &Chen~\etal~\cite{chen2019closerfewshot}& ResNet-18 & 72.66$\pm$0.85 & 89.40$\pm$0.43 \\
\hline
\end{tabular}
}
\vspace{.8mm}
\caption{Performance comparison for FRN under two different class split settings. We include some competitive baselines using the same split as Chen~\etal~\cite{chen2019closerfewshot} for reference. Results of FRN are averaged over 10,000 trials with 95\% confidence interval. $\flat$ denotes the use of non-episodic pre-training. 
}
\label{tab:closer_split}
\end{table}

\subsection{Pre-Training}
We introduce the pre-training technique for FRN in Sec.~\ref{sec:pretrain} and apply it to mini-ImageNet and tiered-ImageNet in Table~\ref{tab:imagenet}. Here, we apply it to FRN with ResNet-12 backbone for un-cropped CUB, as in Sec.~\ref{sec:fine_grained} and Table~\ref{tab:cub_origin}.

For non-episodic FRN pre-training, we run 1200 epochs using SGD with initial learning rate 0.1 and Nesterov momentum 0.9. We cut the learning rate by a factor of 10 at epochs 600 and 900.  Subsequent episodic fine-tuning uses the same optimizer for 600 epochs, but with initial learning rate 1e-3, decreased by a factor of 10 at epochs 300 and 500. For the 5-shot FRN model, we use the standard 20-way episodes. The 1-shot model uses 25-way episodes. 

As shown in Table~\ref{tab:pretrain_cub}, pre-training still improves the final accuracy, but the gain compared to training from scratch becomes much smaller than for mini-ImageNet (see Table~\ref{tab:ablation_pretrain}). This is reasonable, as CUB is only about $\frac{1}{6}$ the size of mini-ImageNet. It is thus comparatively easier in this setting for the FRN trained episodically from scratch to find a good optimum, and more difficult for pre-training to improve.

\begin{table}
\centering
\resizebox{.4\textwidth}{!}{
\scriptsize
\setlength\tabcolsep{5pt}
\hskip-.02\textwidth
\begin{tabular} {  l  c  c}
\hline
\textbf{training setting} & \textbf{1-shot} & \textbf{5-shot} \\
\hline
from scratch & \textbf{83.55$\pm$0.19}  & 92.92$\pm$0.10  \\
\hline
after pre-train & 73.46$\pm$0.20 & 84.20$\pm$0.13 \\

finetune & 83.23$\pm$0.18  & \textbf{93.59$\pm$0.09} \\
\hline
\end{tabular}
}
\vspace{.8mm}
\caption{Impact of pre-training for FRN on CUB using raw images as input. Pre-training can slightly boost the performance compared to training from scratch.}
\label{tab:pretrain_cub}
\end{table}

\section{Additional Visualizations}
\label{sup:visual}
    
    Additional image reconstruction trials as in Fig.~\ref{fig:regenerate} of the main paper begin on the following page.
    
\newpage
    \begin{figure}
        \centering
        \includegraphics[width=\linewidth]{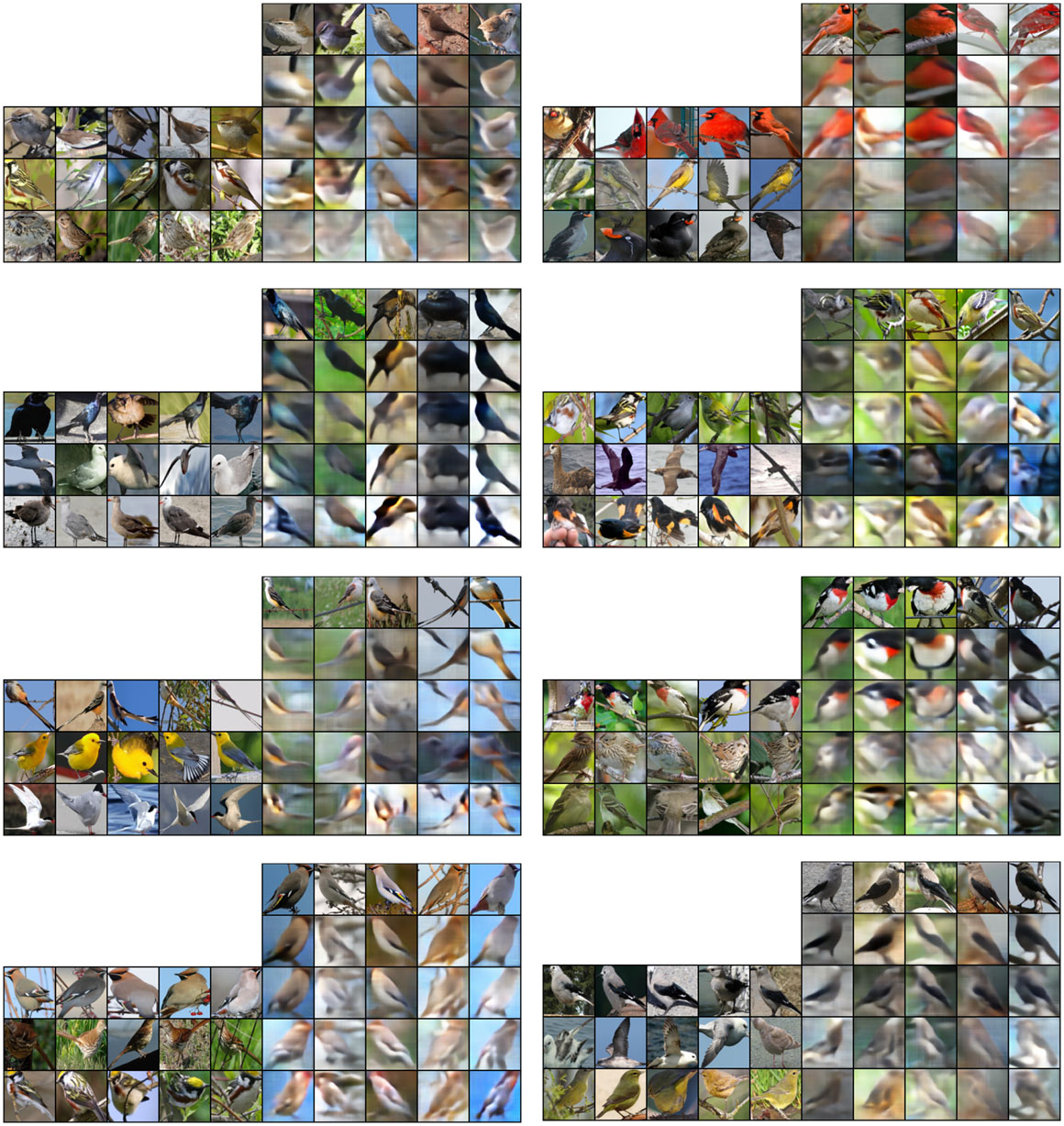}
        \caption{Additional image reconstruction visualizations for CUB. Formatting follows Fig.~\ref{fig:regenerate}: support images are given on the left while target images and reconstructions are on the right. First row images are targets, second row images are autoencoded, third row images are reconstructed from the same class, and fourth and fifth row images are reconstructed from different classes. Same-class reconstructions are clearly superior to those from different classes.}
        \label{fig:cub_supp}
    \end{figure}

\newpage
    \begin{figure}
        \centering
        \includegraphics[width=\linewidth]{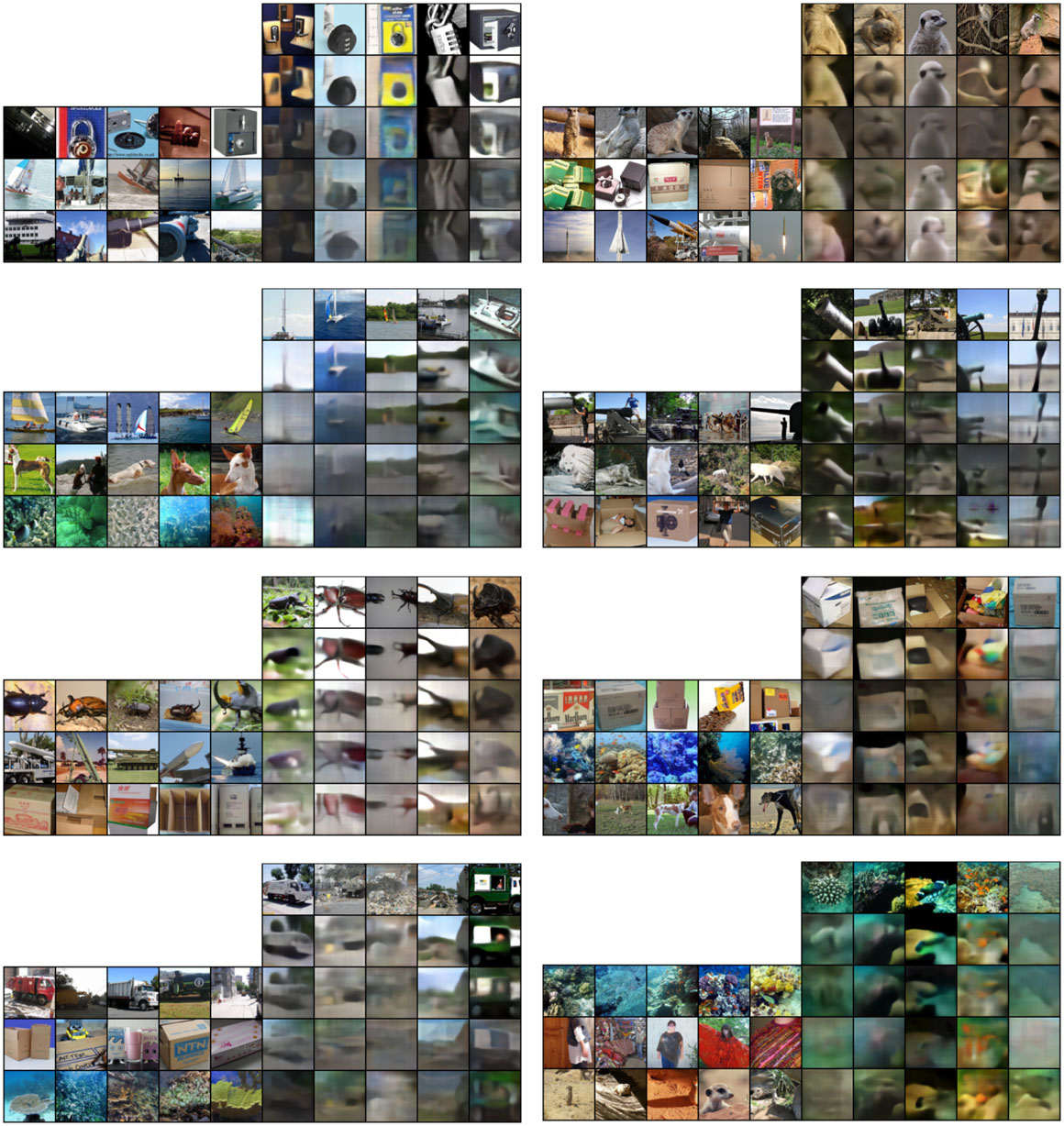}
        \caption{Additional image reconstruction visualizations for mini-ImageNet. Formatting follows Fig.~\ref{fig:regenerate}: support images are given on the left while target images and reconstructions are on the right. First row images are targets, second row images are autoencoded, third row images are reconstructed from the same class, and fourth and fifth row images are reconstructed from different classes. Same-class reconstructions tend to gray out or darken the colors, but are much more faithful shape-wise than those from different classes.}
        \label{fig:min_supp}
    \end{figure}

\end{document}